\documentclass{article}

\PassOptionsToPackage{numbers, compress}{natbib}
\usepackage[final]{neurips_2019}

\usepackage[utf8]{inputenc} 
\usepackage[T1]{fontenc}    
\usepackage{hyperref}       
\usepackage{url}            
\usepackage{booktabs}       
\usepackage{amsfonts}       
\usepackage{nicefrac}       
\usepackage{microtype}      
\usepackage{subcaption}
\usepackage{multirow}
\usepackage{amsmath}
\usepackage{algorithm}
\usepackage{algorithmic}
\usepackage{wrapfig}

\usepackage{tikz}
\usetikzlibrary{calc}
\tikzset{>=latex}

\title{Regularizing Trajectory Optimization\\ with Denoising Autoencoders}

\author{%
  Rinu Boney\thanks{Equal contribution, rest in alphabetical order}\\
  Aalto University \& Curious AI \\
  \texttt{rinu.boney@aalto.fi} \\
  \And
  Norman Di Palo$^*$\\
  Sapienza University of Rome \\
  \texttt{normandipalo@gmail.com} \\
  \And
  Mathias Berglund \\
  Curious AI \\
  \And
  Alexander Ilin \\
  Aalto University \& Curious AI \\
  \And
  Juho Kannala \\
  Aalto University \\
  \And
  Antti Rasmus \\
  Curious AI \\
  \And
  Harri Valpola \\
  Curious AI \\
}

\begin{document}

\maketitle

\begin{abstract}
Trajectory optimization using a learned model of the environment is one of the core elements of model-based reinforcement
learning. This procedure often suffers from exploiting inaccuracies of the learned model. We propose to regularize trajectory optimization by means of a denoising autoencoder that is trained on the same trajectories as the model of the environment. We show that the proposed regularization leads to improved
planning with both gradient-based and gradient-free optimizers.
We also demonstrate that using regularized trajectory optimization leads to rapid initial learning
in a set of popular motor control tasks, which suggests that the proposed approach can be a useful tool
for improving sample efficiency.
\end{abstract}

\section{Introduction}

State-of-the-art reinforcement learning (RL) often requires a large number of interactions with the environment to learn even relatively simple tasks \cite{duan2016benchmarking}.
It is generally believed that model-based RL can provide better sample-efficiency
\cite{deisenroth2013survey, arulkumaran2017brief, chua2018deep} but showing this in practice has been challenging. In this paper, we
propose a way to improve planning in model-based RL and show that it can lead to
improved performance and better sample efficiency.

In model-based RL, planning is done by computing the expected result
of a sequence of future actions using an explicit model of the environment.
Model-based planning has been demonstrated to be efficient in many
applications where the model (a simulator) can be built using first principles.
For example, model-based control is widely used in robotics and has been used to solve
challenging tasks such as human locomotion \cite{tassa2012synthesis, tassa2014control} and dexterous in-hand manipulation \cite{lowrey2018plan}.

In many applications, however, we often do not have the luxury of an accurate simulator
of the environment. Firstly, building even an approximate simulator can be very costly even for processes whose dynamics is well understood. Secondly, it can be challenging to
align the state of an existing simulator with the state of the observed process in order
to plan. Thirdly, the environment is often non-stationary due to, for example, hardware failures in robotics, change of the input feed and deactivation of materials in industrial process control. Thus, learning the model of the environment is the only
viable option in many applications and learning needs to be done for a live system.
And since many real-world systems are very complex, we are likely to need powerful
function approximators, such as deep neural networks, to learn the dynamics of the environment.

However, planning using a learned (and therefore inaccurate) model of the environment
is very difficult in practice. The process of optimizing the sequence of future actions
to maximize the expected return (which we call trajectory optimization) can easily
exploit the inaccuracies of the model and suggest a very unreasonable plan which
produces highly over-optimistic predicted rewards. This optimization process works
similarly to adversarial attacks \cite{akhtar2018threat, huang2017adversarial, szegedy2013intriguing, Dalvi:2004:AC:1014052.1014066} where the input of a trained model is
modified to achieve the desired output. In fact, a more efficient trajectory optimizer is
more likely to fall into this trap. This can arguably be the reason why gradient-based
optimization (which is very efficient at for example learning the models) has not been widely
used for trajectory optimization.

In this paper, we study this adversarial effect of model-based planning in
several environments and show that it poses a problem particularly in high-dimensional control spaces. We also propose to remedy this problem by regularizing
trajectory optimization using a denoising autoencoder (DAE) \cite{vincent2010stacked}.
The DAE is trained to denoise trajectories that appeared in the past experience and
in this way the DAE learns the distribution of the collected trajectories. During trajectory
optimization, we use the denoising error of the DAE as a regularization term
that is subtracted from the maximized objective function. The intuition is that the denoising error will be
large for trajectories that are far from the training distribution, signaling that the dynamics model predictions will be less reliable as it has not been trained on such data. 
Thus, a good trajectory has to give a high predicted return and it can be only moderately novel in the light of past experience.


In the experiments, we demonstrate that the proposed regularization significantly
diminishes the adversarial effect of trajectory optimization with learned models.
We show that the proposed regularization works well with both 
gradient-free and gradient-based optimizers (experiments are done with cross-entropy method \cite{botev2013cross} and Adam \cite{kingma:adam}) in both open-loop and closed-loop control.
We demonstrate that improved trajectory optimization translates to excellent results in early parts of training in standard motor-control
tasks and achieve competitive performance after a handful of interactions with the
environment.


\section{Model-Based Reinforcement Learning}

In this section, we explain the basic setup of model-based RL and present the notation
used. At every time step $t$, the environment is in state $s_t$, the agent performs
action $a_t$, receives reward $r_{t} = r(s_t, a_t)$ and the environment transitions to new state $s_{t+1} = f(s_t, a_t)$. The agent acts based on the observations $o_t=o(s_t)$
which is a function of the environment state.
In a fully observable Markov decision process (MDP), the agent observes
full state $o_t = s_t$. In a partially observable Markov decision process (POMDP),
the observation $o_t$ does not completely reveal $s_t$.
The goal of the agent is select actions $\{a_0, a_1, \ldots \}$ so as to maximize
the return, which is the expected cumulative reward
$\mathbb{E} \left[ \sum_{t=0}^\infty r(s_t, a_t) \right]$.

In the model-based approach, the agent builds the dynamics model
of the environment (forward model).
For a fully observable environment, 
the forward model can be a fully-connected neural network trained to predict the state
transition from time $t$ to $t+1$:
\begin{align}
  s_{t+1} = f_{\theta}(s_t, a_t) \,.
\label{eq:f}
\end{align}
In partially observable environments, the forward model can be a recurrent neural
network trained to directly predict the future observations based on past observations
and actions:
\begin{align}
  o_{t+1} = f_{\theta}(o_0, a_0, \ldots, o_{t}, a_t) \,.
\label{eq:frnn}
\end{align}
In this paper, we assume access to the reward function and that it can be computed from the agent observations, that is $r_{t} = r(o_t, a_t)$.

At each time step $t$, the agent uses the learned forward model to plan the sequence
of future actions $\{a_t, \ldots, a_{t+H}\}$ so as to maximize the expected cumulative future reward. 
\begin{align*}
  G(a_t, \ldots, a_{t+H}) &= \mathbb{E} \left[\sum_{\tau=t}^{t+H} r(o_\tau, a_\tau)\right]
\nonumber \\
  a_t, \dots , a_{t+H} &= \arg \max G(a_t, \ldots, a_{t+H})
\,.
\nonumber
\end{align*}
This process is called trajectory optimization.
The agent uses the learned model of the environment to compute the objective
function $G(a_t, \ldots, a_{t+H})$.
The model \eqref{eq:f} or \eqref{eq:frnn} is unrolled $H$ steps into the future using
the current plan $\{a_t, \ldots, a_{t+H}\}$.

The optimized sequence of actions from trajectory optimization can be directly applied to the environment (open-loop control). It can also be provided as suggestions to a human
operator with the possibility for the human to change the plan (human-in-the-loop).
Open-loop control is challenging because the dynamics model has to be able to make accurate long-range  predictions. An approach which works better in practice
is to take only the first action of the optimized trajectory and then re-plan
at each step (closed-loop control). Thus, in closed-loop control, we account for
possible modeling errors and feedback from the environment. In the control literature, this flavor of model-based RL is called model-predictive control (MPC) \cite{mayne2000constrained, mpcbook, kouvaritakis2001non, nagabandi2018neural}.

The typical sequence of steps performed in model-based RL are: 1)~collect data,
2)~train the forward model $f_\theta$, 3)~interact with the environment using
MPC (this involves trajectory optimization in every time step), 4)~store the data collected during the last interaction and continue to step 2. The
algorithm is outlined in Algorithm~\ref{alg:mbrl}.

\begin{algorithm}[t]
\caption{End-to-end model-based reinforcement learning}
\label{alg:mbrl}
\begin{algorithmic}
\STATE Collect data $\mathbb{D}$ by random policy.
\FOR{each episode}
\STATE Train dynamics model $f_\theta$ using $\mathbb{D}$.
\FOR{time $t$ until the episode is over} 
\STATE Optimize trajectory $\{a_t, o_{t+1}, \ldots, a_{t+H}, o_{t+H+1}\}$.
\STATE Implement the first action $a_t$ and get new observation $o_t$.
\ENDFOR
\STATE Add data $\{(s_1, a_1, \ldots, a_T, o_T)\}$
from the last episode to $\mathbb{D}$.
\ENDFOR

\end{algorithmic}
\end{algorithm}

\section{Regularized Trajectory Optimization}

\subsection{Problem with using learned models for planning}

In this paper, we focus on the inner loop of model-based RL which is trajectory optimization using a \emph{learned}
forward model $f_\theta$. Potential inaccuracies of the trained model cause
substantial difficulties for the planning process. Rather than optimizing what really happens, planning can easily end up exploiting the weaknesses of the predictive model.
Planning is effectively an adversarial attack against the agent's
own forward model. This results in a wide gap between expectations based on the model
and what actually happens.

\begin{figure*}[tp]
\begin{center}
\begin{tabular}{ccc}
\includegraphics[width=.31\textwidth,trim={13mm 12mm 16mm 10mm},clip]{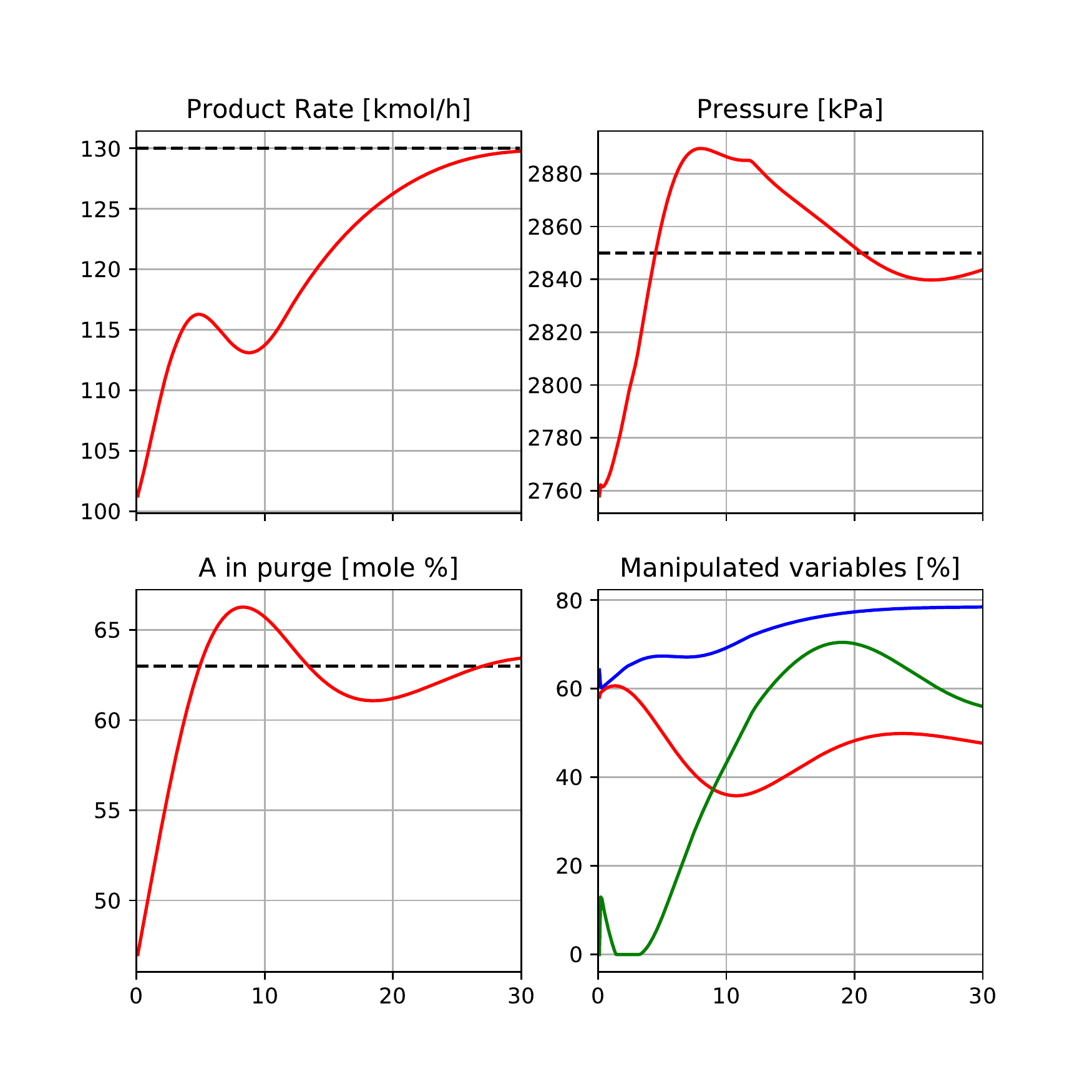}
&
\includegraphics[width=.31\textwidth,trim={12mm 12mm 16mm 10mm},clip]{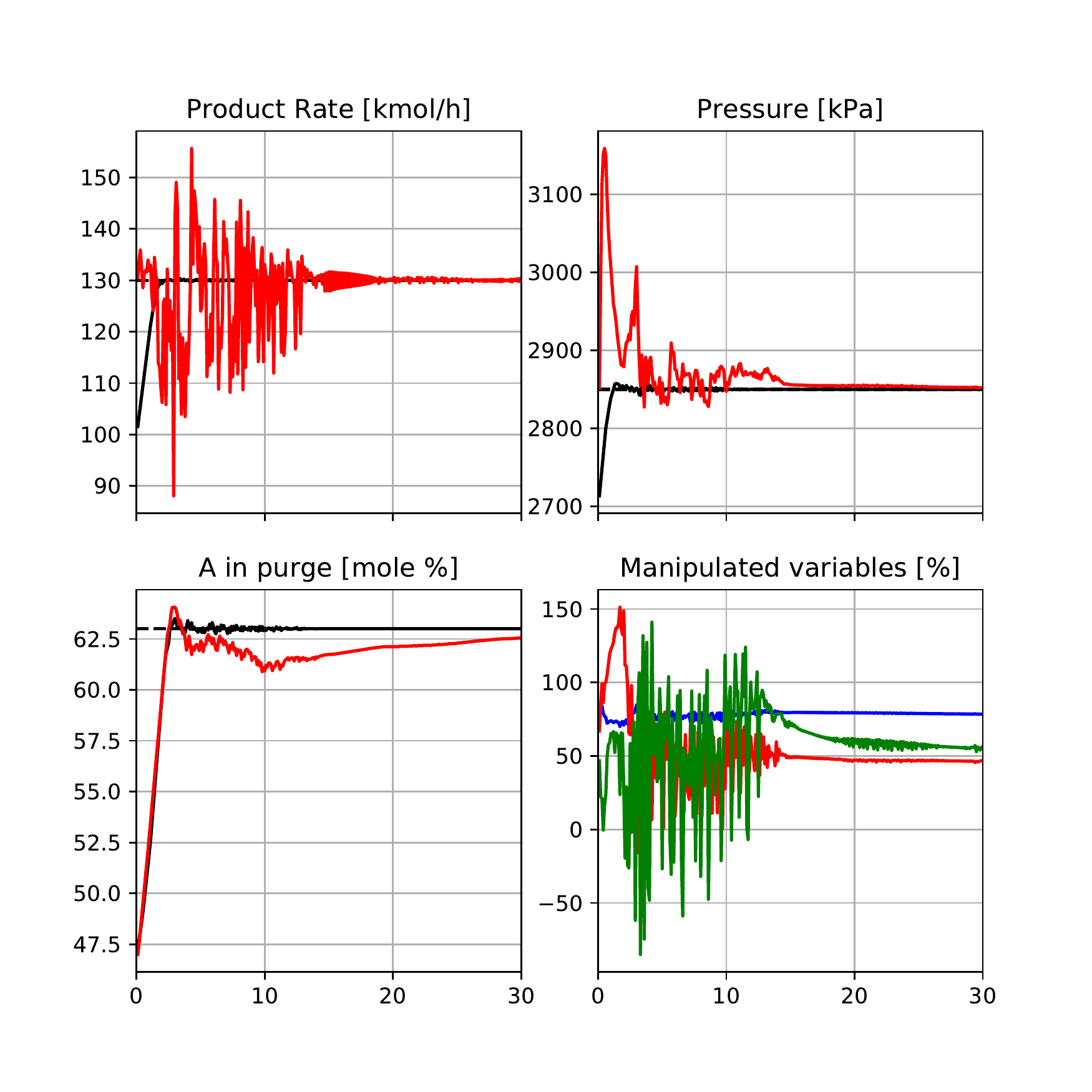}
&
\includegraphics[width=.31\textwidth,trim={12mm 12mm 16mm 10mm},clip]{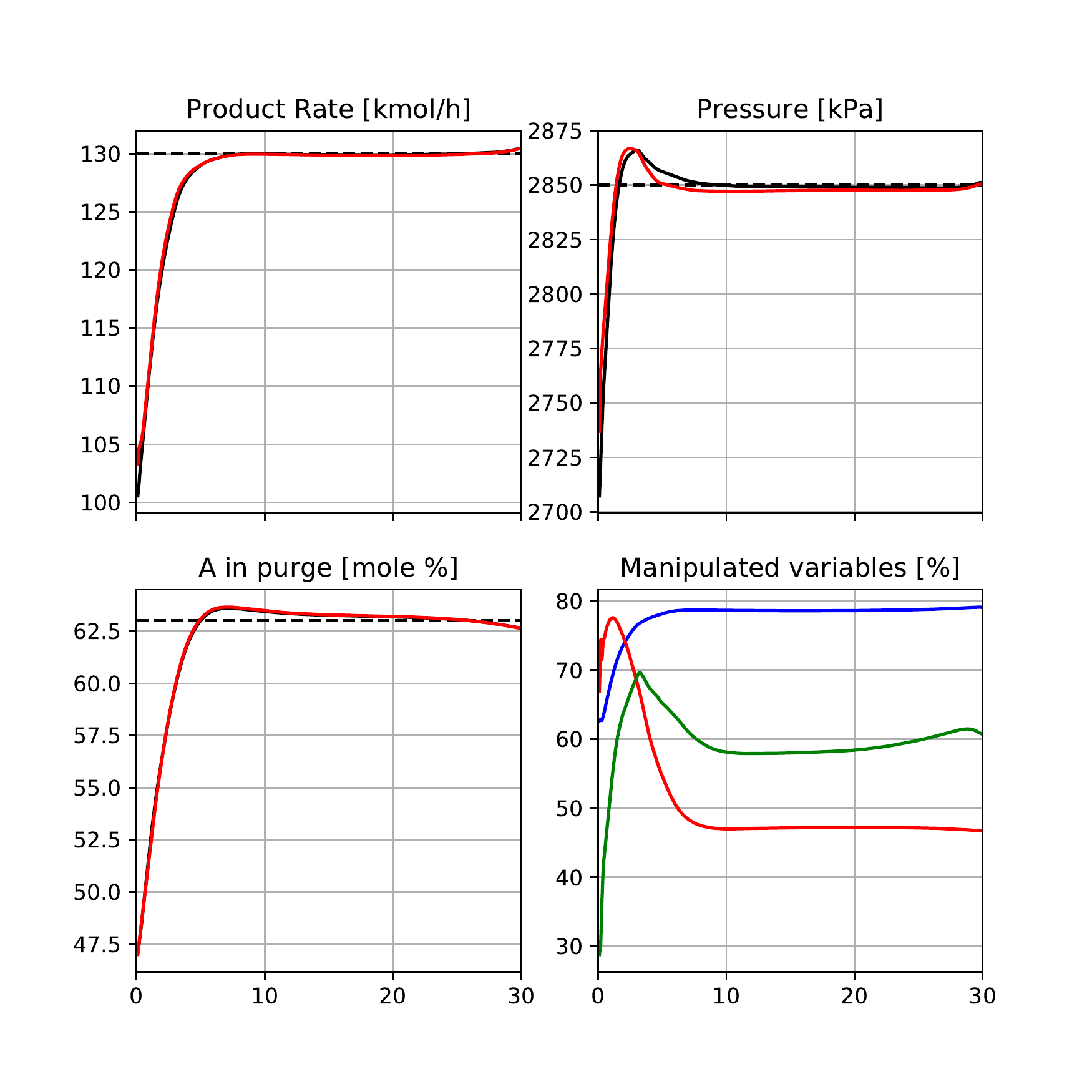}
\\
(a) Multiloop PI control
&
(b) No regularization
&
(c) DAE regularization
\end{tabular}
\end{center}
\caption{Open-loop planning for a continuous nonlinear two-phase reactor from \cite{TE4}.
Three subplots in every subfigure show three measured variables (solid lines): product rate,
pressure and A in the purge. The black curves represent
the model's imagination while the red curves represent the reality if those controls are applied
in an open-loop mode. The targets for the variables are shown with dashed lines.
The fourth (low right) subplots show the three manipulated variables: valve for feed 1 (blue),
valve for feed 2 (red) and valve for stream 3 (green).
}
\label{f:te4}
\end{figure*}

We demonstrate this problem using a simple industrial process control benchmark
from \citep{TE4}. The problem is to control a continuous nonlinear reactor
by manipulating three valves which control flows in two feeds and one output
stream. Further details of the process and the control problem are given in
Appendix~\ref{appendix:ipc}.
The task considered in \citep{TE4} is to change the product rate of the process
from 100 to 130 kmol/h.
Fig.~\ref{f:te4}a shows how this task can be performed using a set of PI controllers
proposed in \citep{TE4}. We trained a forward model of the
process using a recurrent neural network \eqref{eq:frnn} and the data
collected by implementing the PI control strategy for a set of randomly generated
targets. Then we optimized the trajectory for the considered task using 
gradient-based optimization, which produced results in Fig.~\ref{f:te4}b.
One can see that the proposed control signals are changed abruptly and the
trajectory imagined by the model significantly deviates from reality.
For example, the pressure constraint (of max 300 kPa) is violated.
This example demonstrates how planning can easily exploit the weaknesses of the predictive model.

\subsection{Regularizing Trajectory Optimization with Denoising Autoencoders}

We propose to regularize the trajectory optimization with denoising autoencoders (DAE).
The idea is that we want to reward familiar trajectories and penalize unfamiliar ones
because the model is likely to make larger errors for the unfamiliar ones.

This can be achieved by adding a regularization term to the objective function:
\begin{equation}
  G_\text{reg} = G + \alpha \log p(o_t, a_t \ldots, o_{t+H}, a_{t+H}) \,,
\label{eq:Greg_orig}
\end{equation}
where $p(o_t, a_t, \ldots, o_{t+H}, a_{t+H})$ is the probability of observing a given
trajectory in the past experience and $\alpha$ is a tuning hyperparameter.
In practice, instead of using the joint probability of the whole trajectory,
we use marginal probabilities over short windows of size $w$:
\begin{equation}
  G_\text{reg} = G + \alpha \sum_{\tau=t}^{t+H-w} \log p(x_\tau)
\label{eq:Greg}
\end{equation}
where $x_\tau = \{o_\tau, a_\tau, \ldots o_{\tau+w}, a_{\tau+w} \}$
is a short window of the optimized trajectory.

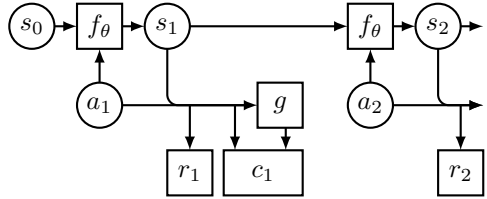
\begin{figure}
\begin{center}
\begin{tikzpicture}[thick, scale=.3]
\usetikzlibrary{shapes.misc}

\newcommand{\fnode}[1]{
	\draw [black] ($(#1)+(1,1)$) rectangle ($(#1)+(1,1)$)
}

\tikzset{cross/.style={cross out, draw=black, minimum size=2*(#1-\pgflinewidth),
inner sep=0pt, outer sep=0pt}, cross/.default={1pt}}

\path[use as bounding box] (0,-3) rectangle (27,8);

\coordinate (x) at (1,1);

\node [] at ($(x)+(1,5)$) {$s_{0}$};
\draw [] ($(x)+(1,5)$) circle [radius=1];
\draw [->] ($(x)+(2,5)$) -- ($(x)+(3,5)$);

\path (x) + (3,0) coordinate (x);

\node [] at ($(x)+(1,5)$) {$f_\theta$};
\draw [black] ($(x)+(0,4)$) rectangle ($(x)+(2,6)$);
\draw [->] ($(x)+(2,5)$) -- ($(x)+(3,5)$);  

\node [] at ($(x)+(1,1.5)$) {$a_1$};
\draw ($(x)+(1,1.5)$) circle [radius=1];
\draw [->] ($(x)+(1,2.5)$) -- ($(x)+(1,4)$);  

\draw [-] ($(x)+(2,1.5)$) -- ($(x)+(7.5,1.5)$);  

\path (x) + (3,0) coordinate (x);

\node [] at ($(x)+(1,5)$) {$s_{1}$};
\draw [] ($(x)+(1,5)$) circle [radius=1];
\draw [->] ($(x)+(2,5)$) -- ($(x)+(9,5)$);  

\draw [->, rounded corners]
    ($(x)+(1,4)$) -- ($(x)+(1,1.5)$) -- ($(x)+(5,1.5)$);  

\path (x) + (3,1.5) coordinate (x);

\node [] at ($(x)+(-1,-3)$) {$r_1$};
\draw [black] ($(x)+(-2,-4)$) rectangle ($(x)+(0,-2)$);
\draw [->] ($(x)+(-1,0)$) -- ($(x)+(-1,-2)$);  

\node [] at ($(x)+(3,0)$) {$g$};
\draw [black] ($(x)+(2,-1)$) rectangle ($(x)+(4,1)$);

\node [] at ($(x)+(2.25,-3)$) {$c_1$};
\draw [black] ($(x)+(0.5,-4)$) rectangle ($(x)+(4,-2)$);
\draw [->] ($(x)+(1,0)$) -- ($(x)+(1,-2)$);  
\draw [->] ($(x)+(3.25,-1)$) -- ($(x)+(3.25,-2)$);  

\path (x) + (1,-1.5) coordinate (x);

\path (x) + (5,0) coordinate (x);

\node [] at ($(x)+(1,5)$) {$f_\theta$};
\draw [black] ($(x)+(0,4)$) rectangle ($(x)+(2,6)$);
\draw [->] ($(x)+(2,5)$) -- ($(x)+(3,5)$);

\node [] at ($(x)+(1,1.5)$) {$a_2$};
\draw ($(x)+(1,1.5)$) circle [radius=1];
\draw [->] ($(x)+(1,2.5)$) -- ($(x)+(1,4)$);  

\draw [->] ($(x)+(2,1.5)$) -- ($(x)+(6,1.5)$);  

\path (x) + (3,0) coordinate (x);

\node [] at ($(x)+(1,5)$) {$s_{2}$};
\draw [] ($(x)+(1,5)$) circle [radius=1];
\draw [->] ($(x)+(2,5)$) -- ($(x)+(3,5)$);

\draw [->, rounded corners, thick]
    ($(x)+(1,4)$) -- ($(x)+(1,1.5)$) -- ($(x)+(3,1.5)$);

\path (x) + (3,1.5) coordinate (x);

\node [] at ($(x)+(-1,-3)$) {$r_2$};
\draw [black] ($(x)+(-2,-4)$) rectangle ($(x)+(0,-2)$);
\draw [->] ($(x)+(-1,0)$) -- ($(x)+(-1,-2)$);  

\end{tikzpicture}
\vspace{-10pt}
\end{center}
\caption{
Example: fragment of a computational graph used during trajectory optimization in an MDP. Here, window size $w=1$,
that is the DAE penalty term is $c_1 = \lVert g([s_1, a_1]) - [s_1, a_1] \rVert^2$.
}
\label{f:graph}
\end{figure}

Suppose we want to find the optimal sequence of actions by maximizing \eqref{eq:Greg}
with a gradient-based optimization procedure. We can compute gradients
$\frac{\partial G_\text{reg}}{\partial a_i}$
by backpropagation in a computational graph where the trained forward model
is unrolled into the future (see Fig.~\ref{f:graph}).
In such backpropagation-through-time procedure, one needs to compute the gradient
with respect to actions $a_i$. 
\begin{equation}
  \frac{\partial G_\text{reg}}{\partial a_i} =
  \frac{\partial G}{\partial a_i}
  + \alpha \sum_{\tau=i}^{i+w} \frac{\partial x_\tau}{\partial a_i} \frac{\partial}{\partial x_\tau} \log p(x_\tau)
\,,
\label{eq:dG}
\end{equation}
where we denote by $x_\tau$ a concatenated vector of observations $o_\tau, \ldots o_{\tau+w}$ and actions $a_\tau, \ldots a_{\tau+w}$, over a window of size $w$.
Thus to enable a regularized gradient-based optimization procedure, we need
means to compute $\frac{\partial}{\partial x_\tau} \log p(x_\tau)$.

In order to evaluate $\log p(x_\tau)$ (or its derivative), one needs
to train a separate model $p(x_\tau)$ of the past experience, which is the task
of unsupervised learning.
In principle, any probabilistic model can be used for that.
In this paper, we propose to regularize trajectory optimization with a denoising autoencoder
(DAE) which does not build an explicit probabilistic model $p(x_\tau)$
but rather learns to approximate the derivative
of the log probability density.
The theory of denoising \cite{miyasawa1961empirical, raphan2011least} states that the optimal
denoising function $g(\tilde x)$ (for zero-mean Gaussian corruption) is given by:
\[
g(\tilde x) = \tilde x + \sigma_n^2
\frac{\partial }{\partial \tilde x} \log p(\tilde x)\,,
\]
where $p(\tilde x)$ is the probability density function for data $\tilde x$ corrupted with noise
and $\sigma_n$ is the standard deviation of the Gaussian corruption.
Thus, the DAE-denoised signal minus the original gives the gradient of the
log-probability of the data distribution convolved with a Gaussian distribution:
$
  \frac{\partial }{\partial \tilde x} \log p(\tilde x)
  \propto g(x) - x
  \,.
$
Assuming $\frac{\partial }{\partial \tilde x} \log p(\tilde x) \approx \frac{\partial }{\partial x} \log p(x)$ yields
\begin{equation}
  \frac{\partial G_\text{reg}}{\partial a_i} =
  \frac{\partial G}{\partial a_i}
  + \alpha \sum_{\tau=i}^{i+w} \frac{\partial x_\tau}{\partial a_i} (g(x_\tau) - x_\tau) \,.
\end{equation}
Using $\frac{\partial }{\partial \tilde x} \log p(\tilde x)$ instead 
of $\frac{\partial }{\partial x} \log p(x)$ can behave better in
practice because it is similar to replacing $p(x)$ with its Parzen window estimate
\cite{vincent2011connection}.
In automatic differentiation software, this gradient can be computed by
adding the penalty term $\lVert g(x_\tau) - x_\tau \rVert^2$ to $G$ and stopping the gradient propagation through $g$. In practice, stopping the gradient through $g$ did not yield any benefits in our experiments compared to simply adding the penalty term
$\lVert g(x_\tau) - x_\tau \rVert^2$ to the cumulative reward, so we used the simple penalty term in our experiments.
Also, this kind of regularization can easily be used with gradient-free optimization methods such as cross-entropy method (CEM) \cite{botev2013cross}.

Our goal is to tackle high-dimensional problems and expressive models of dynamics. Neural networks tend to fare better than many other techniques in modeling high-dimensional distributions. However, using a neural network or any other flexible parameterized model to estimate the input distribution poses a dilemma: the regularizing network which is supposed to keep planning from exploiting the inaccuracies of the dynamics model will itself have weaknesses which planning will then exploit. Clearly, DAE will also have inaccuracies but planning will not exploit them because unlike most other density models, DAE develops an explicit model of the gradient of logarithmic probability density.

The effect of adding DAE regularization in the industrial process control benchmark discussed in the previous section is shown in Fig.~\ref{f:te4}c.

\subsection{Related work}


Several methods have been proposed for planning with learned dynamics models. Locally linear time-varying models \cite{kumar2016optimal, NIPS2014_5444} and Gaussian processes \cite{deisenroth2011pilco, ko2007gaussian} or mixture of Gaussians \cite{rommel2019gaussian} are data-efficient but have problems scaling to high-dimensional environments. Recently, deep neural networks have been successfully applied to model-based RL. \citet{nagabandi2018neural} use deep neural networks as dynamics models in model-predictive control to achieve good performance, and then shows how model-based RL can be fine-tuned with a model-free approach to achieve even better performance. \citet{chua2018deep} introduce PETS, a method to improve model-based performance by estimating and propagating uncertainty with an ensemble of networks and sampling techniques. They demonstrate how their approach can beat several recent model-based and model-free techniques. \citet{clavera2018model} combines model-based RL and meta-learning with MB-MPO, training a policy to quickly adapt to slightly different learned dynamics models, thus enabling faster learning.

\citet{levine2013guided} and \citet{kumar2016optimal} use a KL divergence penalty between action distributions to stay close to the training distribution. Similar bounds are also used to stabilize training of policy gradient methods \cite{schulman2015trust, schulman2017proximal}. While such a KL penalty bounds the evolution of action distributions, the proposed method also bounds the familiarity of states, which could be important in high-dimensional state spaces. While penalizing unfamiliar states also penalize exploration, it allows for more controlled and efficient exploration. Exploration is out of the scope of the paper but was studied in \cite{di2018improving}, where a non-zero optimum of the proposed DAE penalty was used as an intrinsic reward to alternate between familiarity and exploration.

\section{Experiments on Motor Control}

We show the effect of the proposed regularization for control in standard Mujoco environments: Cartpole, Reacher, Pusher, Half-cheetah and Ant available in \cite{brockman2016openai}.
See the description of the environments in Appendix~\ref{sec:env}. We use the Probabilistic Ensembles with Trajectory Sampling (PETS) model from \cite{chua2018deep} as the baseline,
which achieves the best reported results on all the considered tasks except for Ant.
The PETS model consists of an ensemble of probabilistic neural networks and uses particle-based trajectory sampling to regularize trajectory optimization.
We re-implemented the PETS model using the code provided by the authors as a reference.

\subsection{Regularized trajectory optimization with models trained with PETS}
\label{sec:asperf}

\begin{figure}[t]
\begin{center}
\begin{tabular}{cc}
\includegraphics[width=.475\textwidth,trim={5mm 65mm 3mm 3mm},clip]{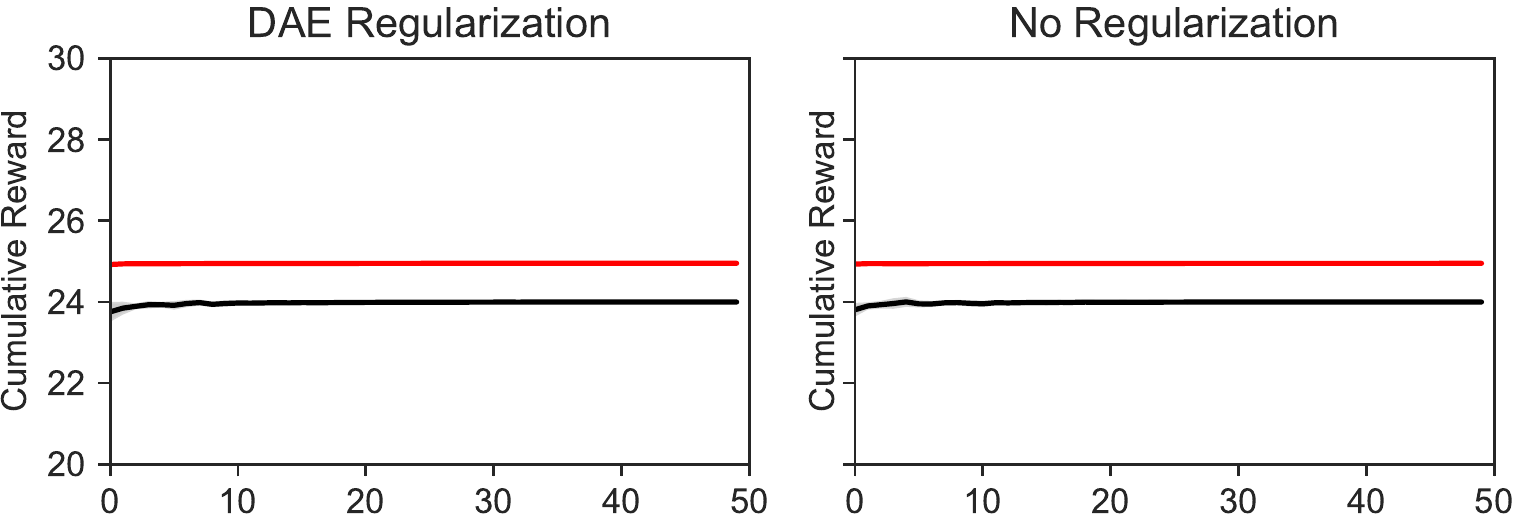}
&
\includegraphics[width=.475\textwidth,trim={5mm 65mm 3mm 3mm},clip]{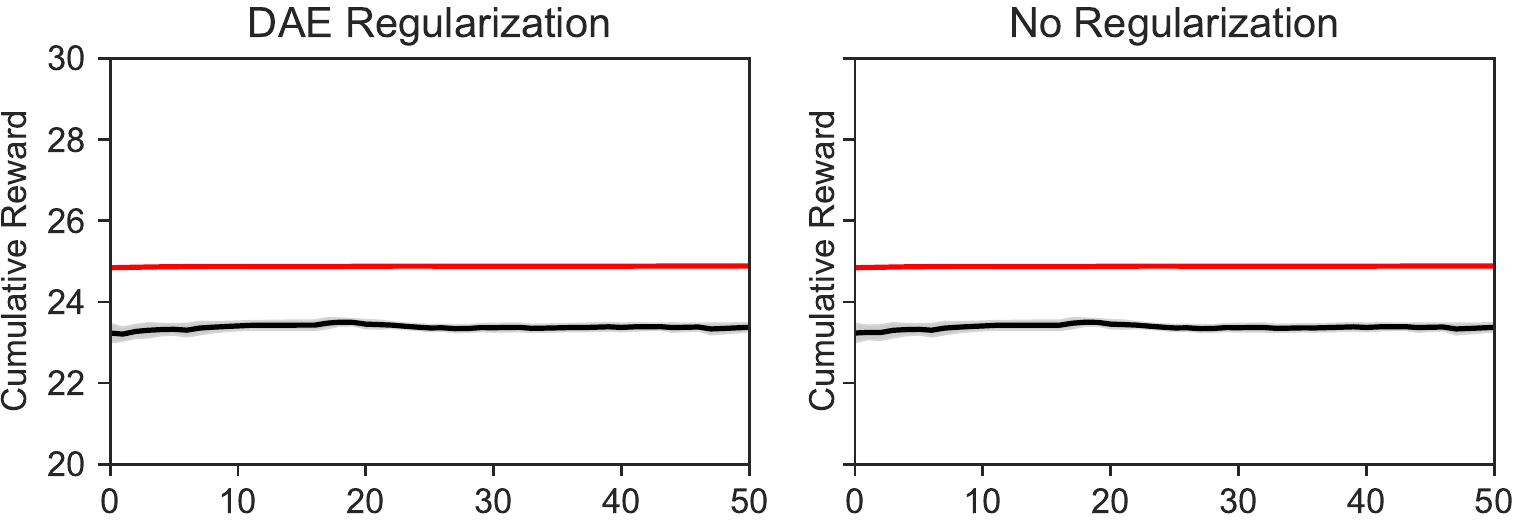}
\\
\includegraphics[width=.475\textwidth,trim={3mm 60mm 3mm 7.8mm},clip]{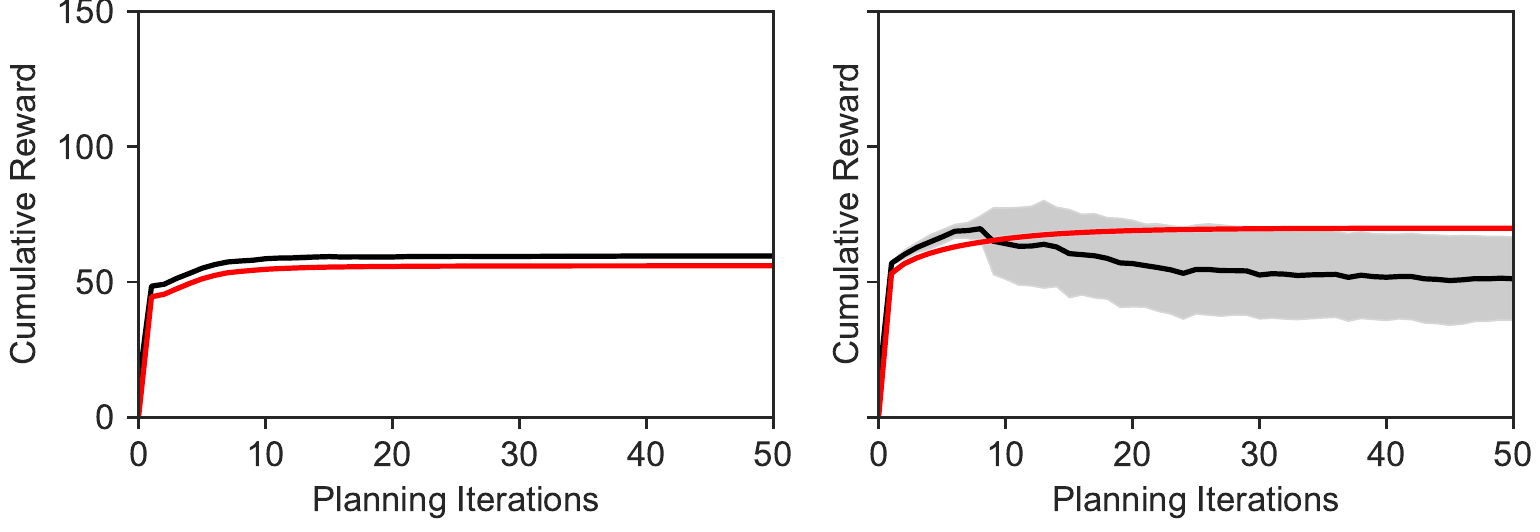}
&
\includegraphics[width=.475\textwidth,trim={10mm 60mm 3mm 7.8mm},clip]{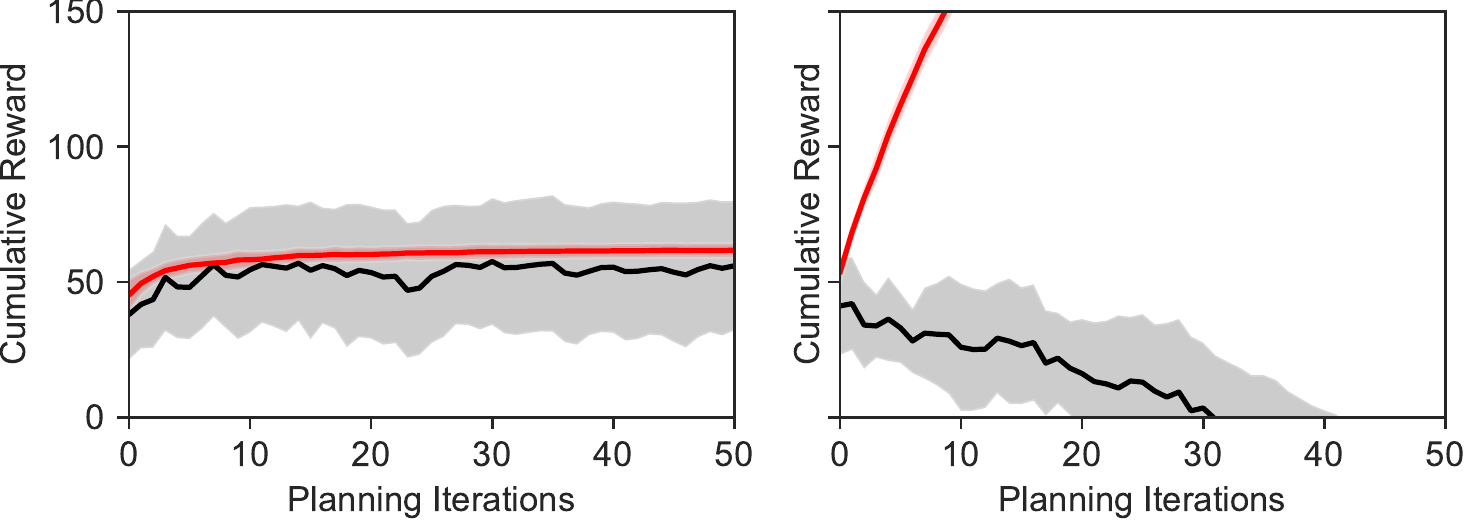}
\\
Trajectory optimization with CEM
&
Trajectory optimization with Adam
\\
\end{tabular}
\end{center}
\caption{Visualization of trajectory optimization at timestep $t = 50$. Each row has the same model
but a different optimization method. The models are obtained by 5 episodes of end-to-end training.
Row above: Cartpole environment. Row below: Half-cheetah environment.
Here, the red lines denote the rewards predicted by the model (imagination) and the black lines
denote the true rewards obtained when applying the sequence of optimized actions (reality).
For a low-dimensional action space (Cartpole), trajectory optimizers do not
exploit inaccuracies of the dynamics model and hence DAE regularization does not affect the performance noticeably.
For a higher-dimensional action space (Half-cheetah), gradient-based optimization without any
regularization easily exploits inaccuracies of the dynamics model but DAE regularization is
able to prevent this. The effect is less pronounced with gradient-free optimization but still
noticeable.
}
\label{f:cp_effect}
\end{figure}

In MPC, the innermost loop is open-loop control which is then turned to closed-loop control by
taking in new observations and replanning after each action.
Fig.~\ref{f:cp_effect} illustrates the adversarial effect during open-loop trajectory optimization and
how DAE regularization mitigates it. In Cartpole environment, the learned model is very good already after
a few episodes of data and trajectory optimization stays within the data distribution. As there is no
problem to begin with, regularization does not improve the results. In Half-cheetah environment,
trajectory optimization manages to exploit the inaccuracies of the model which is particularly
apparent in gradient-based Adam. DAE regularization improves both but the effect is much stronger
with Adam.

The problem is exacerbated in closed-loop control since it continues optimization from the solution achieved in the previous time step, effectively iterating more per action.
We demonstrate how regularization can improve closed-loop trajectory optimization in the Half-cheetah environment.
We first train three PETS models for 300 episodes using the best hyperparameters reported in \cite{chua2018deep}. We then evaluate the performance of the three models on five episodes using
four different trajectory optimizers: 1)~Cross-entropy method (CEM) which was used during training
of the PETS models, 2)~Adam, 3)~CEM with the DAE regularization and 4)~Adam with the DAE regularization.
The results averaged across the three models and the five episodes are presented in
Table~\ref{t:closed-loop-pets}.

\begin{table*}[tbh]
  \caption{Comparison of PETS with CEM and Adam optimizers in Half-cheetah}
  \label{t:closed-loop-pets}
  \centering
  \begin{tabular}{lcccc}
    \toprule
    Optimizer & CEM & CEM + DAE & Adam & Adam + DAE \\
    \midrule
    Average Return & $10955 \pm 2865$ & $12967 \pm 3216$ & -- & $12796 \pm 2716$ \\
    \bottomrule
  \end{tabular}
\end{table*}

We first note that planning with Adam fails completely without regularization: the proposed actions
lead to unstable states of the simulator. Using Adam with the DAE regularization fixes this problem
and the obtained results are better than the CEM method originally used in PETS. 
CEM appears to regularize trajectory optimization but not as efficiently CEM+DAE. These open-loop
results are consistent with the closed-loop results in Fig.~\ref{f:cp_effect}.

\subsection{End-to-end training with regularized trajectory optimization}
\label{sec:e2e}

In the following experiments, we study the performance of end-to-end training with different
trajectory optimizers used during training. 
Our agent learns according to the algorithm presented in Algorithm~\ref{alg:mbrl}. Since the environments
are fully observable, we use a feedforward neural network as in \eqref{eq:f} to model the dynamics of the
environment. Unlike PETS, we did not use an ensemble of probabilistic networks as the forward model.
We use a single probabilistic network which predicts the mean and variance of the next state (assuming a Gaussian distribution) given the current state and action. Although we only use the mean prediction, we found that also training to predict the variance improves the stability of the training.

For all environments, we use a dynamics model with the same architecture: three hidden layers of size 200 with the Swish non-linearity \cite{ramachandran2017swish}. Similar to prior works, we train the dynamics model to predict the difference between $s_{t+1}$ and $s_t$ instead of predicting $s_{t+1}$ directly.
We train the dynamics model for 100 or more epochs (see Appendix~\ref{appendix:details})
after every episode. This is a larger number of updates compared to five epochs used in
\cite{chua2018deep}. We found that an increased number of updates has a large effect on the performance
for a single probabilistic model and not so large effect for the ensemble of models used in PETS.
This effect is shown in Fig.~\ref{f:ablation}.

For the denoising autoencoder, we use the same architecture as the dynamics model.
The state-action pairs in the past episodes were corrupted with zero-mean Gaussian noise and the DAE was trained to denoise it. Important hyperparameters used in our experiments are reported in the Appendix~\ref{appendix:details}. For DAE-regularized trajectory optimization we used either CEM or
Adam as optimizers.

\begin{figure}[t]
\begin{center}
\includegraphics[width=\textwidth]{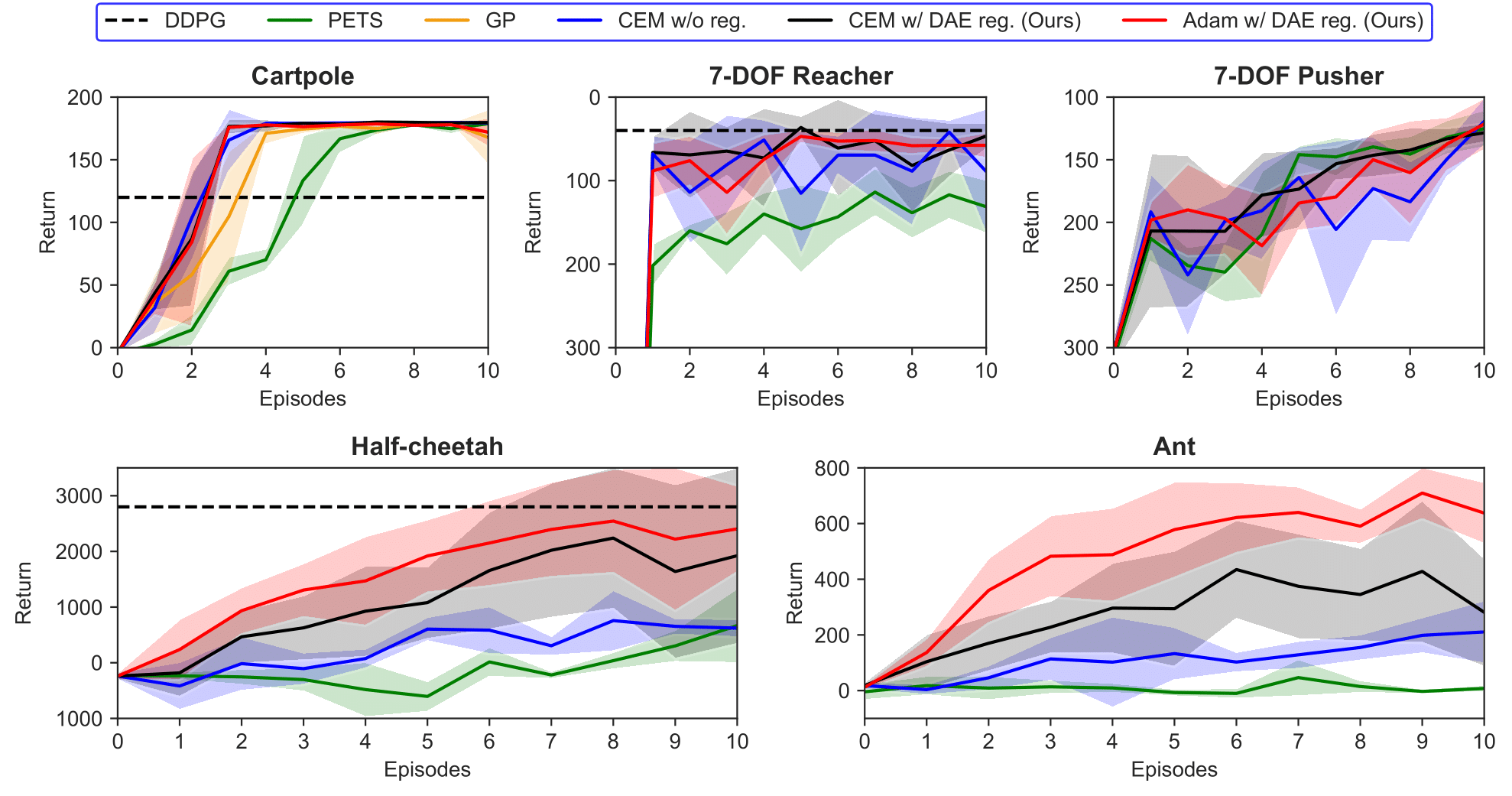}
\end{center}
\caption{Results of our experiments on the five benchmark environments, in comparison to PETS \cite{chua2018deep}. We show the return obtained in each episode. All the results are averaged across 5 seeds, with the shaded area representing standard deviation. PETS is a recent state-of-the-art model-based RL algorithm and GP-based (Gaussian Processes) control algorithms are well known to be sample-efficient and are extensively used for the control of simple systems.}
\label{f:results}
\end{figure}

The learning progress of the compared algorithms is presented in Fig.~\ref{f:results}.
Note that we report the \emph{average} returns across different seeds, not
the maximum return seen so far as was done in \cite{chua2018deep}.\footnote{
Because of the different metric used, the PETS
results presented in this paper may appear worse than in \cite{chua2018deep}. However,
we verified that our implementation of PETS obtains similar results to \cite{chua2018deep}
for the metric used in \cite{chua2018deep}.}
In Cartpole, all the methods converge to the maximum cumulative reward but
the proposed method converges the fastest.
In the Cartpole environment, we also compare
to a method which uses Gaussian Processes (GP) as the dynamics model (algorithm denoted
GP-E in \cite{chua2018deep}, which considers only the expectation of the next state prediction). The implementation of the GP algorithm was obtained from the code provided by \cite{chua2018deep}.
Interestingly, our algorithm also surpasses the Gaussian Process (GP) baseline, which is known to be a sample-efficient method widely used for control of simple systems.
In Reacher, the proposed method converges to the same asymptotic performance as PETS, but faster. In Pusher, all algorithms perform similarly.

In Half-cheetah and Ant, the proposed method shows very good sample efficiency and very rapid initial learning.
The agent learns an effective running gait in only a couple of episodes.\footnote{
Videos of our agents during training can be found at \url{https://sites.google.com/view/regularizing-mbrl-with-dae/home}.}
The results demonstrate that denoising regularization is effective for both gradient-free and
gradient-based planning, with gradient-based planning performing the best. 
The proposed algorithm learns faster than PETS in the initial phase of training.
It also achieves performance that is competitive with popular model-free algorithms such
as DDPG, as reported in \cite{chua2018deep}.

However, the performance of the proposed method does not improve after
initial 10 episodes, so it does not reach the asymptotic performance of PETS (see results
for PETS for Half-cheetah after 300 episodes in Table~\ref{t:closed-loop-pets}).
This result is evidence of the importance of exploration:
the DAE regularization essentially penalizes exploration, which can harm asymptotic
performance in complex environments.
In PETS, CEM leaves some noise in the trajectories, which might help to obtain better asymptotic performance.
The result presented in Appendix~\ref{appendix:exploration} provides some evidence that at
least a part of the problem is lack of exploration.


\begin{figure}
\centering
\includegraphics[width=.45\textwidth, trim=23 5 45 28, clip]{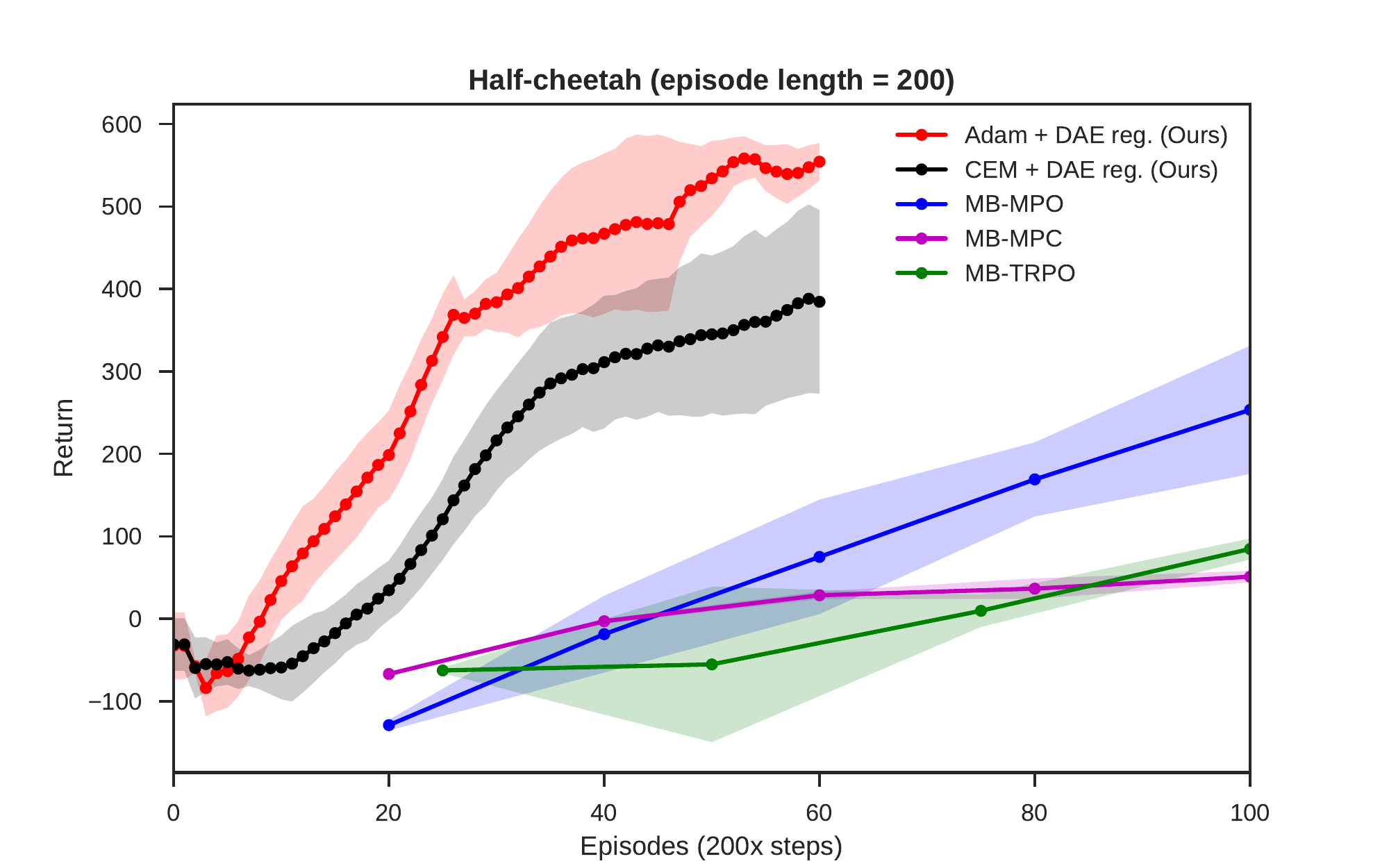}
\caption{Comparison to MB-MPO \cite{clavera2018model}, MB-TRPO \cite{kurutach2018model} and MB-MPC \cite{nagabandi2018neural}
on Half-cheetah. We plot the average return over the last 20 episodes. Our results are averaged across 3 seeds, with the shaded area representing standard deviation. Note that the comparison numbers are picked from \cite{clavera2018model} and the results from the first 20 episodes are not reported.}
\label{f:results_mbmpo}
\end{figure}

We also compare the performance of our method with Model-Based Meta Policy Optimization (MB-MPO) \cite{clavera2018model}, an approach that combines the benefits of model-based RL and meta learning: the algorithm trains a policy using simulations generated by an ensemble of models, learned from data. Meta-learning allows this policy to quickly adapt to the various dynamics, hence learning how to quickly adapt in the real environment, using Model-Agnostic Meta Learning (MAML) \cite{finn2017model}. In Fig.~\ref{f:results_mbmpo} we compare our method to MB-MPO and other model-based methods
included in \cite{clavera2018model}. This experiment is done in the Half-cheetah environment with shorter
episodes (200 timesteps) in order to compare to the results reported in \cite{clavera2018model}.
The results show that our method learns faster than MB-MPO.



\section{Discussion}

In recent years, a lot of effort has been put in making deep reinforcement algorithms more sample-efficient, and thus adaptable to real world scenarios. Model-based reinforcement learning has shown promising results, obtaining sample-efficiency even orders of magnitude better than model-free counterparts, but these methods have often suffered from sub-optimal performance due to many reasons. As already noted in the recent literature \cite{nagabandi2018neural, chua2018deep}, out-of-distribution errors and model overfitting are often sources of performance degradation when using complex function approximators.
In this work we demonstrated how to tackle this problem using regularized trajectory optimization. 
Our experiments demonstrate that the proposed solution can improve the performance of model-based reinforcement learning.

While trajectory optimization is a key component in model-based RL, there are clearly several other
issues which need to be tackled in complex environments:

\begin{itemize}
\item Local minima for trajectory optimization. There can be multiple trajectories that are reasonable solutions but in-between trajectories can be very bad. For example, we can take a step with a right or left foot but both will not work. We tackled this issue by trying multiple initializations, which worked for the considered environments, but better techniques will be needed for more complex environments.

\item The planning horizon problem. In the presented experiments, the planning procedure did not care
about what happens after the planning horizon. This was not a problem for the considered environments due to nicely formatted reward. Other solutions like value functions, multiple time scales or hierarchy for planning are required with sparser reward problems. All of these are compatible with model-based RL.

\item Open-loop vs.\ closed-loop (compounding errors).
The implicit planning assumption of trajectory optimization is open-loop control.
However, MPC only takes the first action and then replans (closed-loop control). If the outcome is uncertain (e.g., due to stochastic environments or imperfect forward model),
this can lead to overly pessimistic controls.

\item Local optima of the policy. This is the well-known exploration-exploitation dilemma.
If the model has never seen data of alternative trajectories, it may predict their consequences incorrectly and never try them (because in-between trajectories can be genuinely worse).
Good trajectory optimization (exploitation) can harm long-term performance because it reduces exploration, but we believe that it is better to add explicit exploration. With model-based RL, intrinsically motivated exploration is a particularly interesting option because it is possible to balance exploration and the expected cost. This is particularly important in hazardous environments where safe exploration is needed.

\item High-dimensional input space. Sensory systems like cameras, lidars and microphones can produce vast amounts of data and it is infeasible to plan based on detailed prediction on low level such as pixels. Also, predictive models of pixels may miss the relevant state.

\item Changing environments. All the considered environments were static but real-world systems keep changing. Online learning and similar techniques are needed to keep track of the changing environment.

\end{itemize}

Still, model-based RL is an attractive approach and
not only due to its sample-efficiency. Compared
to model-free approaches, model-based learning makes safe exploration and adding known constraints or first-principles models much easier. We believe that the proposed method can be a viable solution for real-world control tasks especially where safe exploration is of high importance.

We are currently working on applying the proposed methods for real-world problems such as assisting operators of complex industrial processes and for control of autonomous mobile machines.


\subsubsection*{Acknowledgments}

We would like to thank Jussi Sainio, Jari Rosti and Isabeau Pr\'emont-Schwarz for their valuable contributions in the experiments on industrial process control.

\bibliography{neurips_2019}

\begin{thebibliography}{37}
\providecommand{\natexlab}[1]{#1}
\providecommand{\url}[1]{\texttt{#1}}
\expandafter\ifx\csname urlstyle\endcsname\relax
  \providecommand{\doi}[1]{doi: #1}\else
  \providecommand{\doi}{doi: \begingroup \urlstyle{rm}\Url}\fi

\bibitem[Akhtar and Mian(2018)]{akhtar2018threat}
Naveed Akhtar and Ajmal Mian.
\newblock Threat of adversarial attacks on deep learning in computer vision: A
  survey.
\newblock \emph{IEEE Access}, 6:\penalty0 14410--14430, 2018.

\bibitem[Arulkumaran et~al.(2017)Arulkumaran, Deisenroth, Brundage, and
  Bharath]{arulkumaran2017brief}
Kai Arulkumaran, Marc~Peter Deisenroth, Miles Brundage, and Anil~Anthony
  Bharath.
\newblock A brief survey of deep reinforcement learning.
\newblock \emph{arXiv preprint arXiv:1708.05866}, 2017.

\bibitem[Botev et~al.(2013)Botev, Kroese, Rubinstein, and
  L'Ecuyer]{botev2013cross}
Zdravko~I Botev, Dirk~P Kroese, Reuven~Y Rubinstein, and Pierre L'Ecuyer.
\newblock The cross-entropy method for optimization.
\newblock In \emph{Handbook of statistics}, volume~31, pages 35--59. Elsevier,
  2013.

\bibitem[Brockman et~al.(2016)Brockman, Cheung, Pettersson, Schneider,
  Schulman, Tang, and Zaremba]{brockman2016openai}
Greg Brockman, Vicki Cheung, Ludwig Pettersson, Jonas Schneider, John Schulman,
  Jie Tang, and Wojciech Zaremba.
\newblock {OpenAI} gym.
\newblock \emph{arXiv preprint arXiv:1606.01540}, 2016.

\bibitem[Chua et~al.(2018)Chua, Calandra, McAllister, and Levine]{chua2018deep}
Kurtland Chua, Roberto Calandra, Rowan McAllister, and Sergey Levine.
\newblock Deep reinforcement learning in a handful of trials using
  probabilistic dynamics models.
\newblock In \emph{Advances in Neural Information Processing Systems 31}, pages
  4759--4770. 2018.

\bibitem[Clavera et~al.(2018)Clavera, Rothfuss, Schulman, Fujita, Asfour, and
  Abbeel]{clavera2018model}
Ignasi Clavera, Jonas Rothfuss, John Schulman, Yasuhiro Fujita, Tamim Asfour,
  and Pieter Abbeel.
\newblock Model-based reinforcement learning via meta-policy optimization.
\newblock \emph{arXiv preprint arXiv:1809.05214}, 2018.

\bibitem[Dalvi et~al.(2004)Dalvi, Domingos, Mausam, Sanghai, and
  Verma]{Dalvi:2004:AC:1014052.1014066}
Nilesh Dalvi, Pedro Domingos, Mausam, Sumit Sanghai, and Deepak Verma.
\newblock Adversarial classification.
\newblock In \emph{Proceedings of the Tenth ACM SIGKDD International Conference
  on Knowledge Discovery and Data Mining}, KDD '04, pages 99--108, 2004.

\bibitem[Deisenroth and Rasmussen(2011)]{deisenroth2011pilco}
Marc Deisenroth and Carl~E Rasmussen.
\newblock {PILCO}: A model-based and data-efficient approach to policy search.
\newblock In \emph{Proceedings of the 28th International Conference on Machine
  Learning (ICML)}, pages 465--472, 2011.

\bibitem[Deisenroth et~al.(2013)Deisenroth, Neumann, Peters,
  et~al.]{deisenroth2013survey}
Marc~Peter Deisenroth, Gerhard Neumann, Jan Peters, et~al.
\newblock A survey on policy search for robotics.
\newblock \emph{Foundations and Trends in Robotics}, 2\penalty0
  (1--2):\penalty0 1--142, 2013.

\bibitem[Di~Palo and Valpola(2018)]{di2018improving}
Norman Di~Palo and Harri Valpola.
\newblock Improving model-based control and active exploration with
  reconstruction uncertainty optimization.
\newblock \emph{arXiv preprint arXiv:1812.03955}, 2018.

\bibitem[Duan et~al.(2016)Duan, Chen, Houthooft, Schulman, and
  Abbeel]{duan2016benchmarking}
Yan Duan, Xi~Chen, Rein Houthooft, John Schulman, and Pieter Abbeel.
\newblock Benchmarking deep reinforcement learning for continuous control.
\newblock In \emph{Proceedings of the 33rd International Conference on Machine
  Learning (ICML)}, pages 1329--1338, 2016.

\bibitem[Finn et~al.(2017)Finn, Abbeel, and Levine]{finn2017model}
Chelsea Finn, Pieter Abbeel, and Sergey Levine.
\newblock Model-agnostic meta-learning for fast adaptation of deep networks.
\newblock In \emph{Proceedings of the 34th International Conference on Machine
  Learning (ICML)}, pages 1126--1135, 2017.

\bibitem[Huang et~al.(2017)Huang, Papernot, Goodfellow, Duan, and
  Abbeel]{huang2017adversarial}
Sandy Huang, Nicolas Papernot, Ian Goodfellow, Yan Duan, and Pieter Abbeel.
\newblock Adversarial attacks on neural network policies.
\newblock \emph{arXiv preprint arXiv:1702.02284}, 2017.

\bibitem[Kingma and Ba(2015)]{kingma:adam}
Diederick~P Kingma and Jimmy Ba.
\newblock Adam: A method for stochastic optimization.
\newblock In \emph{International Conference on Learning Representations}, 2015.

\bibitem[Ko et~al.(2007)Ko, Klein, Fox, and Haehnel]{ko2007gaussian}
Jonathan Ko, Daniel~J Klein, Dieter Fox, and Dirk Haehnel.
\newblock Gaussian processes and reinforcement learning for identification and
  control of an autonomous blimp.
\newblock In \emph{Robotics and Automation, 2007 IEEE International Conference
  on}, pages 742--747. IEEE, 2007.

\bibitem[Kouvaritakis and Cannon(2001)]{kouvaritakis2001non}
Basil Kouvaritakis and Mark Cannon.
\newblock \emph{Non-linear Predictive Control: theory and practice}.
\newblock Iet, 2001.

\bibitem[Kumar et~al.(2016)Kumar, Todorov, and Levine]{kumar2016optimal}
Vikash Kumar, Emanuel Todorov, and Sergey Levine.
\newblock Optimal control with learned local models: Application to dexterous
  manipulation.
\newblock In \emph{Robotics and Automation (ICRA), 2016 IEEE International
  Conference on}, pages 378--383. IEEE, 2016.

\bibitem[Kurutach et~al.(2018)Kurutach, Clavera, Duan, Tamar, and
  Abbeel]{kurutach2018model}
Thanard Kurutach, Ignasi Clavera, Yan Duan, Aviv Tamar, and Pieter Abbeel.
\newblock Model-ensemble trust-region policy optimization.
\newblock In \emph{International Conference on Learning Representations}, 2018.

\bibitem[Levine and Abbeel(2014)]{NIPS2014_5444}
Sergey Levine and Pieter Abbeel.
\newblock Learning neural network policies with guided policy search under
  unknown dynamics.
\newblock In Z.~Ghahramani, M.~Welling, C.~Cortes, N.~D. Lawrence, and K.~Q.
  Weinberger, editors, \emph{Advances in Neural Information Processing Systems
  27}, pages 1071--1079. 2014.

\bibitem[Levine and Koltun(2013)]{levine2013guided}
Sergey Levine and Vladlen Koltun.
\newblock Guided policy search.
\newblock In \emph{Proceedings of the 30th International Conference on Machine
  Learning (ICML)}, pages 1--9, 2013.

\bibitem[Lowrey et~al.(2018)Lowrey, Rajeswaran, Kakade, Todorov, and
  Mordatch]{lowrey2018plan}
Kendall Lowrey, Aravind Rajeswaran, Sham Kakade, Emanuel Todorov, and Igor
  Mordatch.
\newblock Plan online, learn offline: Efficient learning and exploration via
  model-based control.
\newblock \emph{arXiv preprint arXiv:1811.01848}, 2018.

\bibitem[Mayne et~al.(2000)Mayne, Rawlings, Rao, and
  Scokaert]{mayne2000constrained}
David~Q Mayne, James~B Rawlings, Christopher~V Rao, and Pierre~OM Scokaert.
\newblock Constrained model predictive control: Stability and optimality.
\newblock \emph{Automatica}, 36\penalty0 (6):\penalty0 789--814, 2000.

\bibitem[Miyasawa(1961)]{miyasawa1961empirical}
K~Miyasawa.
\newblock An empirical bayes estimator of the mean of a normal population.
\newblock \emph{Bulletin of the International Statistical Institute},
  38\penalty0 (181-188):\penalty0 1--2, 1961.

\bibitem[Nagabandi et~al.(2018)Nagabandi, Kahn, Fearing, and
  Levine]{nagabandi2018neural}
Anusha Nagabandi, Gregory Kahn, Ronald~S Fearing, and Sergey Levine.
\newblock Neural network dynamics for model-based deep reinforcement learning
  with model-free fine-tuning.
\newblock In \emph{2018 IEEE International Conference on Robotics and
  Automation (ICRA)}, pages 7559--7566. IEEE, 2018.

\bibitem[Pong* et~al.(2018)Pong*, Gu*, Dalal, and Levine]{pong*2018temporal}
Vitchyr Pong*, Shixiang Gu*, Murtaza Dalal, and Sergey Levine.
\newblock Temporal difference models: Model-free deep {RL} for model-based
  control.
\newblock In \emph{International Conference on Learning Representations}, 2018.

\bibitem[Ramachandran et~al.(2017)Ramachandran, Zoph, and
  Le]{ramachandran2017swish}
Prajit Ramachandran, Barret Zoph, and Quoc~V Le.
\newblock Searching for activation functions.
\newblock \emph{arXiv preprint arXiv:1710.05941}, 2017.

\bibitem[Raphan and Simoncelli(2011)]{raphan2011least}
Martin Raphan and Eero~P Simoncelli.
\newblock Least squares estimation without priors or supervision.
\newblock \emph{Neural computation}, 23\penalty0 (2):\penalty0 374--420, 2011.

\bibitem[Ricker(1993)]{TE4}
N~Lawrence Ricker.
\newblock Model predictive control of a continuous, nonlinear, two-phase
  reactor.
\newblock \emph{Journal of Process Control}, 3\penalty0 (2):\penalty0 109--123,
  1993.

\bibitem[Rommel et~al.(2019)Rommel, Bonnans, Martinon, and
  Gregorutti]{rommel2019gaussian}
C{\'e}dric Rommel, Fr{\'e}d{\'e}ric Bonnans, Pierre Martinon, and Baptiste
  Gregorutti.
\newblock Gaussian mixture penalty for trajectory optimization problems.
\newblock \emph{Journal of Guidance, Control, and Dynamics}, pages 1--6, 2019.

\bibitem[Rossiter(2003)]{mpcbook}
John Rossiter.
\newblock \emph{Model-based Predictive Control-a Practical Approach}.
\newblock CRC Press, 01 2003.

\bibitem[Schulman et~al.(2015)Schulman, Levine, Moritz, Jordan, and
  Abbeel]{schulman2015trust}
John Schulman, Sergey Levine, Philipp Moritz, Michael Jordan, and Pieter
  Abbeel.
\newblock Trust region policy optimization.
\newblock In \emph{Proceedings of the 32nd International Conference on Machine
  Learning (ICML)}, pages 1889--1897, 2015.

\bibitem[Schulman et~al.(2017)Schulman, Wolski, Dhariwal, Radford, and
  Klimov]{schulman2017proximal}
John Schulman, Filip Wolski, Prafulla Dhariwal, Alec Radford, and Oleg Klimov.
\newblock Proximal policy optimization algorithms.
\newblock \emph{arXiv preprint arXiv:1707.06347}, 2017.

\bibitem[Szegedy et~al.(2013)Szegedy, Zaremba, Sutskever, Bruna, Erhan,
  Goodfellow, and Fergus]{szegedy2013intriguing}
Christian Szegedy, Wojciech Zaremba, Ilya Sutskever, Joan Bruna, Dumitru Erhan,
  Ian Goodfellow, and Rob Fergus.
\newblock Intriguing properties of neural networks.
\newblock \emph{arXiv preprint arXiv:1312.6199}, 2013.

\bibitem[Tassa et~al.(2012)Tassa, Erez, and Todorov]{tassa2012synthesis}
Yuval Tassa, Tom Erez, and Emanuel Todorov.
\newblock Synthesis and stabilization of complex behaviors through online
  trajectory optimization.
\newblock In \emph{2012 IEEE/RSJ International Conference on Intelligent Robots
  and Systems}, pages 4906--4913. IEEE, 2012.

\bibitem[Tassa et~al.(2014)Tassa, Mansard, and Todorov]{tassa2014control}
Yuval Tassa, Nicolas Mansard, and Emo Todorov.
\newblock Control-limited differential dynamic programming.
\newblock In \emph{2014 IEEE International Conference on Robotics and
  Automation (ICRA)}, pages 1168--1175. IEEE, 2014.

\bibitem[Vincent(2011)]{vincent2011connection}
Pascal Vincent.
\newblock A connection between score matching and denoising autoencoders.
\newblock \emph{Neural computation}, 23\penalty0 (7):\penalty0 1661--1674,
  2011.

\bibitem[Vincent et~al.(2010)Vincent, Larochelle, Lajoie, Bengio, and
  Manzagol]{vincent2010stacked}
Pascal Vincent, Hugo Larochelle, Isabelle Lajoie, Yoshua Bengio, and
  Pierre-Antoine Manzagol.
\newblock Stacked denoising autoencoders: Learning useful representations in a
  deep network with a local denoising criterion.
\newblock \emph{Journal of machine learning research}, 11\penalty0
  (Dec):\penalty0 3371--3408, 2010.

\end{thebibliography}

\clearpage

\appendix

\section{Industrial Process Control Benchmark}
\label{appendix:ipc}

To study trajectory optimization, we first consider the problem of control of a simple industrial process. An effective industrial control system could achieve better production and economic efficiency than manually operated controls. In this paper, we learn the dynamics of an industrial process and use it to optimize the controls, by minimizing a cost function. 
In some critical processes, safety is of utmost importance and regularization methods could prevent adaptive control methods from exploring unsafe trajectories.

We consider the problem of control of a continuous nonlinear two-phase reactor from \citep{TE4}. The simulated industrial process
consists of a single vessel that represents a combination of the reactor and separation system.
The process has two feeds: one contains substances A, B and C and the other one is pure A.
Reaction $\text{A}+\text{C} \rightarrow \text{D}$ occurs in the vapour phase. The liquid is
pure D which is the product. The process is manipulated by three valves which regulate the flows
in the two feeds and an output stream which contains A, B and C.
The plant has ten measured variables including the flow rates of the four streams ($F_1, \ldots, F_4$), pressure, liquid holdup volume and mole \% of A, B and C in the purge. The control problem is to transition to a specified product rate and maintain it by manipulating the three valves. The pressure must be kept below the shutdown limit of 3000 kPa. The original paper suggests a multiloop control strategy with several PI controllers \citep{TE4}.

We collected simulated data corresponding to about 0.5M steps of operation by randomly generating control setpoints and using the original multiloop control strategy. The collected data were used to train a neural network model with one layer of 80 LSTM units and a linear readout layer to predict the next-step measurements. The inputs were the three controls and the ten process measurements. The data were pre-processed by scaling such that the standard deviation of the derivatives of each measured variable was of the same scale. This way, the model learned better the dynamics of slow changing variables. We used a fully-connected network architecture with 8 hidden layers (100-200-100-20-100-200-100) to train a DAE on windows of five successive measurement-control pairs. The scaled measurement-control pairs in a window were concatenated to a single vector and corrupted with zero-mean Gaussian noise ($\sigma=0.03$) and the DAE was trained to denoise it.

The trained model was then used for optimizing a sequence of actions to ramp production
as rapidly as possible from $F_4=100$ to $F_4=130$ kmol h$^{-1}$, while satisfying all other
constraints \citep[Scenario II from][]{TE4}. We formulated the objective function as the Euclidean distance to the desired targets (after pre-processing).
The targets corresponded to the following targets for three measurements: $F_4 = 130$~kmol~h$^{-1}$ for product rate,
2850~kPa for pressure and 63~mole~\% for A in the purge.

We optimized a plan of actions 30 hours ahead (or 300 discretized time steps).
The optimized sequence of controls were initialized
with the original multiloop policy applied to the trained dynamics model. That control sequence
together with the predicted and the real outcomes (black and red curves respectively) are shown in Fig.~\ref{f:te4}a.
We then optimized the control sequence using 10000 iterations of Adam with learning rate 0.01
without and with DAE regularization
(with penalty $\alpha \lVert g(x_t) - x_t \rVert^2$).

The results are shown in Fig.~\ref{f:te4}. One can see that without
regularization the control signals are changed abruptly and the trajectory imagined
by the model deviates from reality (Fig.~\ref{f:te4}b).
In contrast, the open-loop plan found with the DAE regularization is noticeably the best solution (Fig.~\ref{f:te4}c), leading the plant to the specified product rate much faster than the human-engineered multiloop PI control from \cite{TE4}. The imagined trajectory (black) stays close to predictions and the targets are reached in about ten hours.
This shows that even in a low-dimensional environment with a large amount of training data, regularization is necessary for planning using a learned model.

\section{Description of Environments}
\label{sec:env}

\textbf{Cartpole}. This task involves a pole attached to a moving cart in a frictionless track, with the goal of swinging up the pole and balancing it in an upright position in the center of the screen. The cost at every time step is measured as the angular distance between the tip of the pole and the target position. Each episode is 200 steps long.

\textbf{Reacher}. This environment consists of a simulated PR2 robot arm with seven degrees of freedom, with the goal of reaching a particular position in space. The cost at every time step is measured as the distance between the arm and the target position. The target position changes every episode. Each episode is 150 steps long.

\textbf{Pusher}. This environment also consists of a simulated PR2 robot arm, with a goal of pushing an object to a target position that changes every episode. The cost at every time step is measured as the distance between the object and the target position. Each episode is 150 steps long.

\textbf{Half-cheetah}. This environment involves training a two-legged "half-cheetah" to run forward as fast as possible by applying torques to 6 different joints. The cost at every time step is measured as the
negative forward velocity. Each episode is 1000 steps long, but the length is reduced to 200 for the benchmark with \cite{clavera2018model}.

\textbf{Ant}. This is the most challenging environment we consider. It consists of a four-legged "ant" controlled by applying torques to its 8 joints. Similar to \cite{pong*2018temporal}, we use a gear ratio to 30 for all joints (this prevents the ant from flipping over frequently during the initially phase of training). The cost, similar to Half-cheetah, is the negative forward velocity. Each episode is 1000 steps long.

\begin{table*}[tbh]
  \caption{Dimensionalities of observation and action spaces of the environments used in this paper}
  \label{t:env-dims}
  \centering
  \begin{tabular}{lcc}
    \toprule
    Environment & Observation space & Action space \\
    \midrule
    Cartpole & 5 & 1 \\
    \midrule
    Reacher & 17 & 7 \\
    \midrule
    Pusher & 20 & 7 \\
    \midrule
    Half-cheetah & 19 & 6 \\
    \midrule
    Ant & 111 & 8 \\ 
    \bottomrule
  \end{tabular}
\end{table*}

\section{Additional Experimental Details}
\label{appendix:details}

For MPC, we use the same planning horizon as PETS (Table~\ref{t:planning-horizon}). The important hyperparameters for all our experiments are shown in Tables \ref{t:hyperparams-pets} and \ref{t:hyperparams-mpo}. We found the DAE noise level, regularization penalty weight $\alpha$ and Adam learning rate to be the most important hyperparameters.

\begin{table*}[!t]
\caption{Important hyperparameters used in our experiments for comparison with PETS. Additionally, for the experiments with gradient-based trajectory optimization on Reacher and Pusher, we initialize the trajectory with a few iterations (2 iterations for Reacher and 5 iterations for Pusher) of CEM.}
\label{t:hyperparams-pets}
\begin{center}
\begin{tabular}{lcccccc}
\toprule
Environment & Optimizer & Optim Iters & Epochs & Adam LR & $\alpha$ & DAE noise $\sigma$ \\
\midrule
\multirow{2}{*}{Cartpole} 
& CEM & 5 & 500 & - & 0.001 & 0.1 \\
& Adam & 10 & 500 & 0.001 & 0.001 & 0.2 \\
\midrule
\multirow{2}{*}{Reacher}
 & CEM & 5 & 500 & - & 0.01 & 0.1 \\
 & Adam & 5 & 300 & 1 & 0.01 & 0.1 \\
\midrule
\multirow{2}{*}{Pusher}
 & CEM & 5 & 500 & - & 0.01 & 0.1 \\
 & Adam & 5 & 300 & 1 & 0.01 & 0.1 \\
\midrule
\multirow{2}{*}{Half-cheetah}
 & CEM & 5 & 100 & - & 2 & 0.1 \\
 & Adam & 10 & 200 & 0.1 & 1 & 0.2 \\
 \midrule
\multirow{2}{*}{Ant}
 & CEM & 5 & 400 & - & 0.045 & 0.3 \\
 & Adam & 10 & 1000 & 0.075 & 0.03 & 0.4 \\
\bottomrule
\end{tabular}
\end{center}
\end{table*}

\begin{table*}[!t]
\caption{Important hyperparameters used in our experiments for comparison with MB-MPO}
\label{t:hyperparams-mpo}
\begin{center}
\begin{tabular}{lccccccc}
\toprule
Environment & Optimizer & Optim Iters & Epochs & Adam LR & $\alpha$ & DAE noise $\sigma$ \\
\midrule
\multirow{2}{*}{Half-cheetah}
 & CEM & 5 & 20 & - & 2 & 0.2 \\
 & Adam & 10 & 40 & 0.1 & 1 & 0.1 \\
\bottomrule
\end{tabular}
\end{center}
\end{table*}

\begin{table*}[!t]
\caption{MPC planning horizons used in our experiments}
\label{t:planning-horizon}
\begin{center}
\begin{tabular}{lccccc}
\toprule
Environment & Cartpole & Reacher & Pusher & Half-cheetah & Ant \\
\midrule
Planning Horizon & 25 & 25 & 25 & 30 & 35 \\
\bottomrule
\end{tabular}
\end{center}
\end{table*}

\begin{figure}[ht]
\centering
\includegraphics[width=.6\textwidth, trim=16 5 45 28, clip]{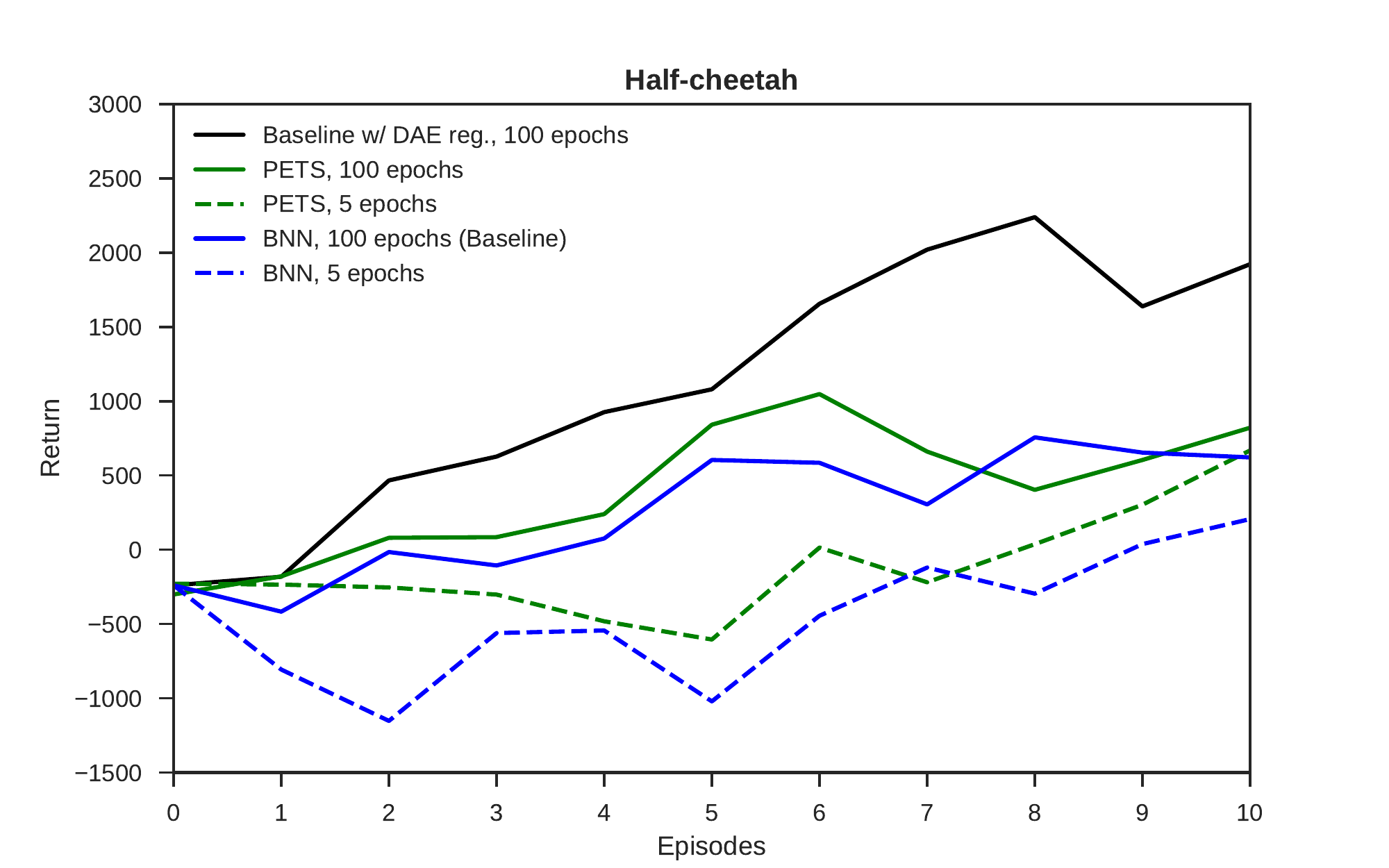}
\caption{Effect of increased number of training epochs after every episode: we can see that training the dynamics model for more epochs after each episode leads to a much better performance in the initial episodes. With this modification, a single dynamics model with no regularization seems to work almost as well as PETS. It can also be clearly seen that the use of denoising regularization enables an improvement in the learning progress. To compare with PETS, we used the CEM optimizer in this ablation study.}
\label{f:ablation}
\end{figure}

\section{Comparison to Gaussian regularization}

\begin{figure}[t]
\begin{center}
\includegraphics[width=0.6\textwidth, trim=16 5 45 28, clip]{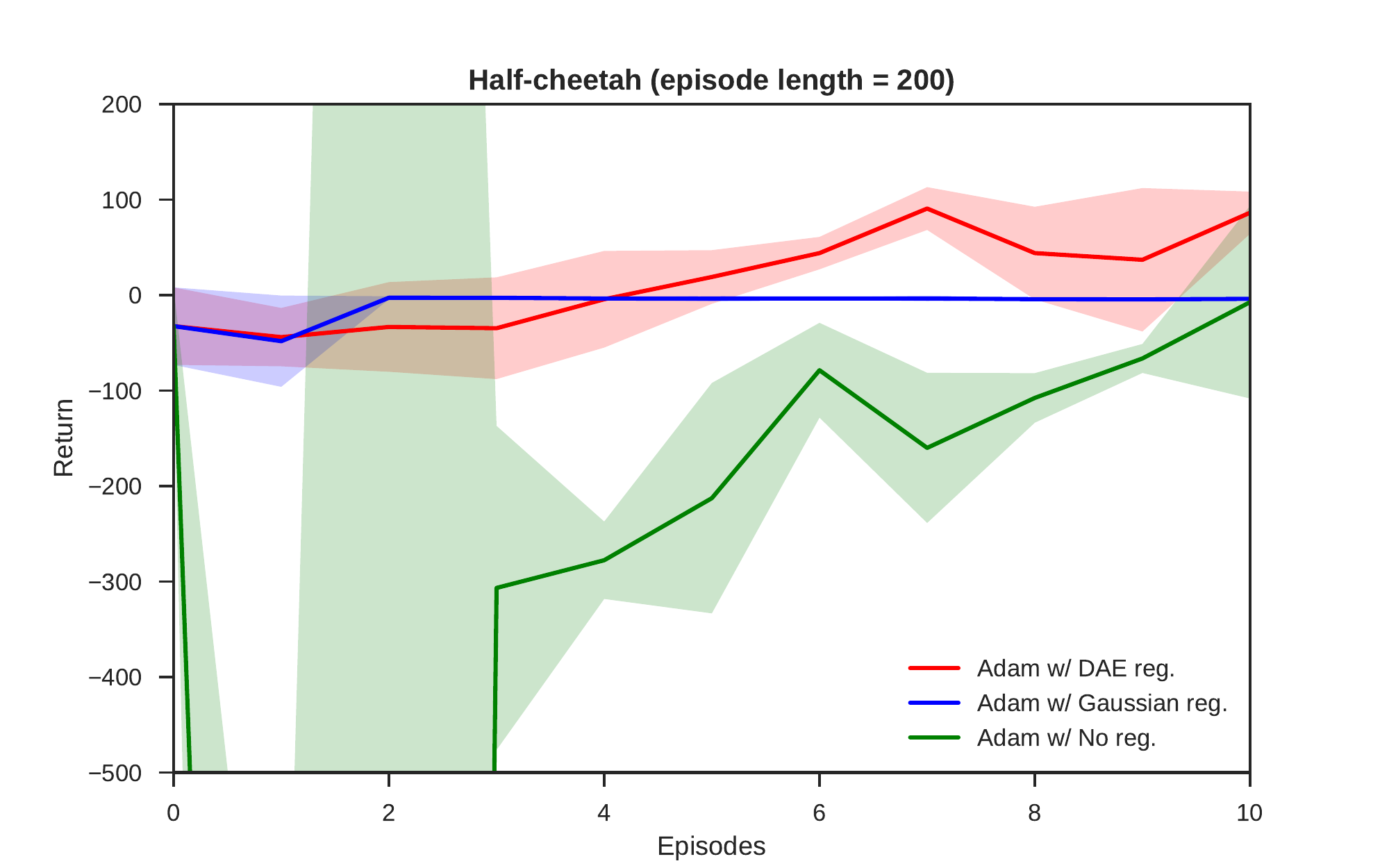}
\caption{Comparison to Gaussian regularization: we can see that trajectory optimization with Adam without any regularization is very unstable and completely fails in the initial episodes. While Gaussian regularization helps in the first few episodes, it is not able to fit the data properly and seems to consistently lead the optimization to a local minimum. As shown earlier in Fig.~\ref{f:results_mbmpo}, denoising regularization is able to successfully regularize the optimization, enabling good asymptotic performance from very few episodes of interaction.}
\label{f:gaussian}
\end{center}
\end{figure}

To emphasize the importance of denoising regularization, we also compare against a simple Gaussian regularization baseline: we fit a Gaussian distribution (with diagonal covariance matrix) to the states and actions in the replay buffer and regularize the trajectory optimization by adding a penalty term to the cost, proportional to the negative log probability of the states and actions in the trajectory (Equation~\ref{eq:Greg}). The performance of this baseline in the Half-cheetah task (with an episode length of 200) is shown in Fig.~\ref{f:gaussian}. We observe that the Gaussian distribution poorly fits the trajectories and consistently leads the optimization to a bad local minimum.

\section{Preliminary Experiments on Exploration}
\label{appendix:exploration}

\begin{figure}[t]
\begin{center}
\includegraphics[width=0.7\textwidth, trim=8 5 40 28, clip]{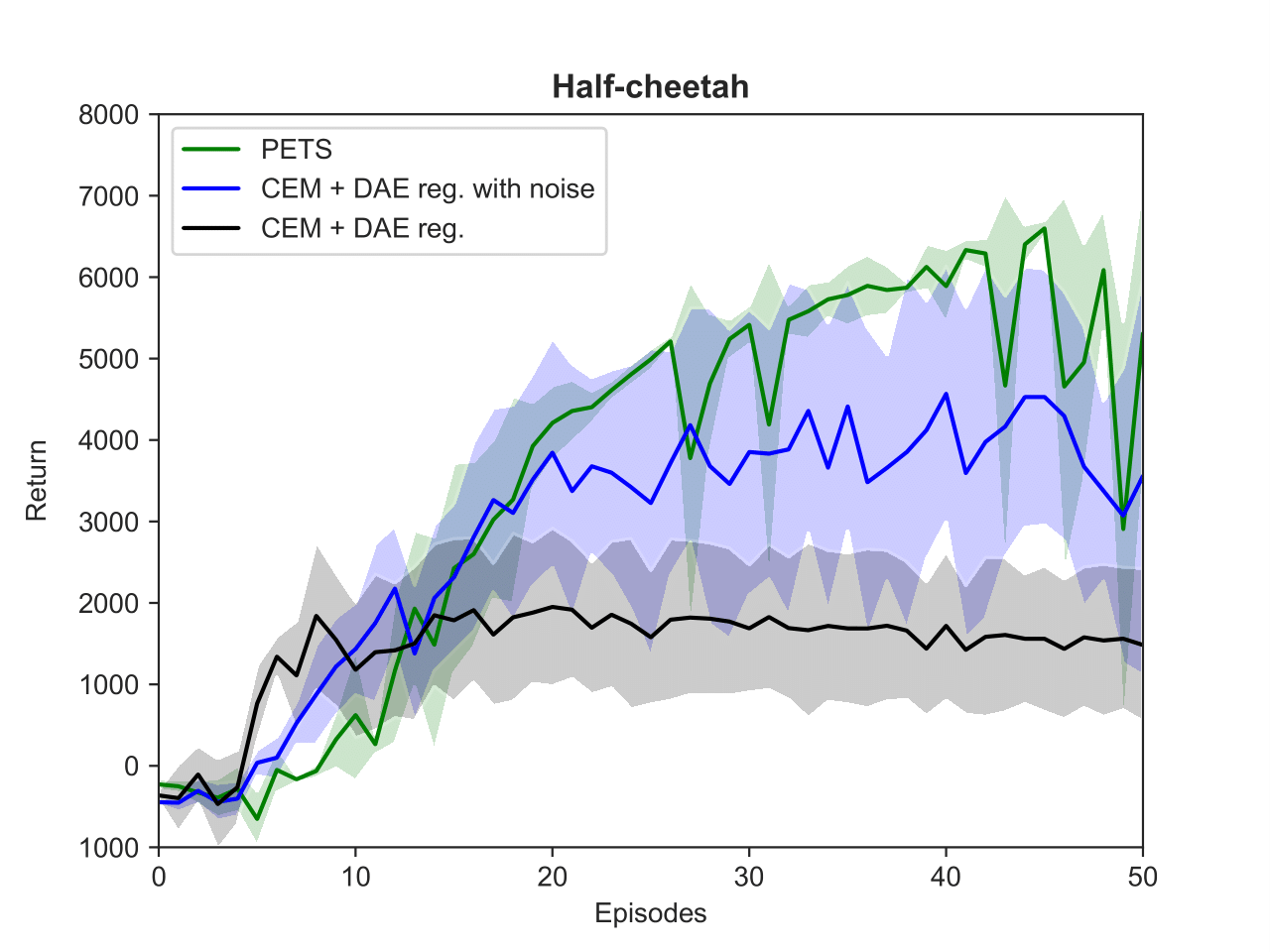}
\caption{In this plot we show the cumulative reward obtained during training by our method when we inject noise to actions in order to improve exploration of the state-action space. Plots are averaged over 5 seeds, and show mean and standard deviation.}
\label{f:hc-noise}
\end{center}
\end{figure}

To improve the asymptotic performance of our agent, we perform some preliminary experiments on exploration by injecting random noise into the optimized actions. In Figure~\ref{f:hc-noise}, we show that asymptotic performance can greatly benefit from random exploration, suggesting a line of future work.

\section{Visualization of Trajectory Optimization in End-to-End Experiments}

\begin{figure*}[tp]
\begin{center}
\begin{tabular}{cc}
Optimizer: CEM & Optimizer: Adam
\\
\addlinespace[3mm]

\includegraphics[width=.5\textwidth,trim={3mm 3mm 3mm 3mm},clip]{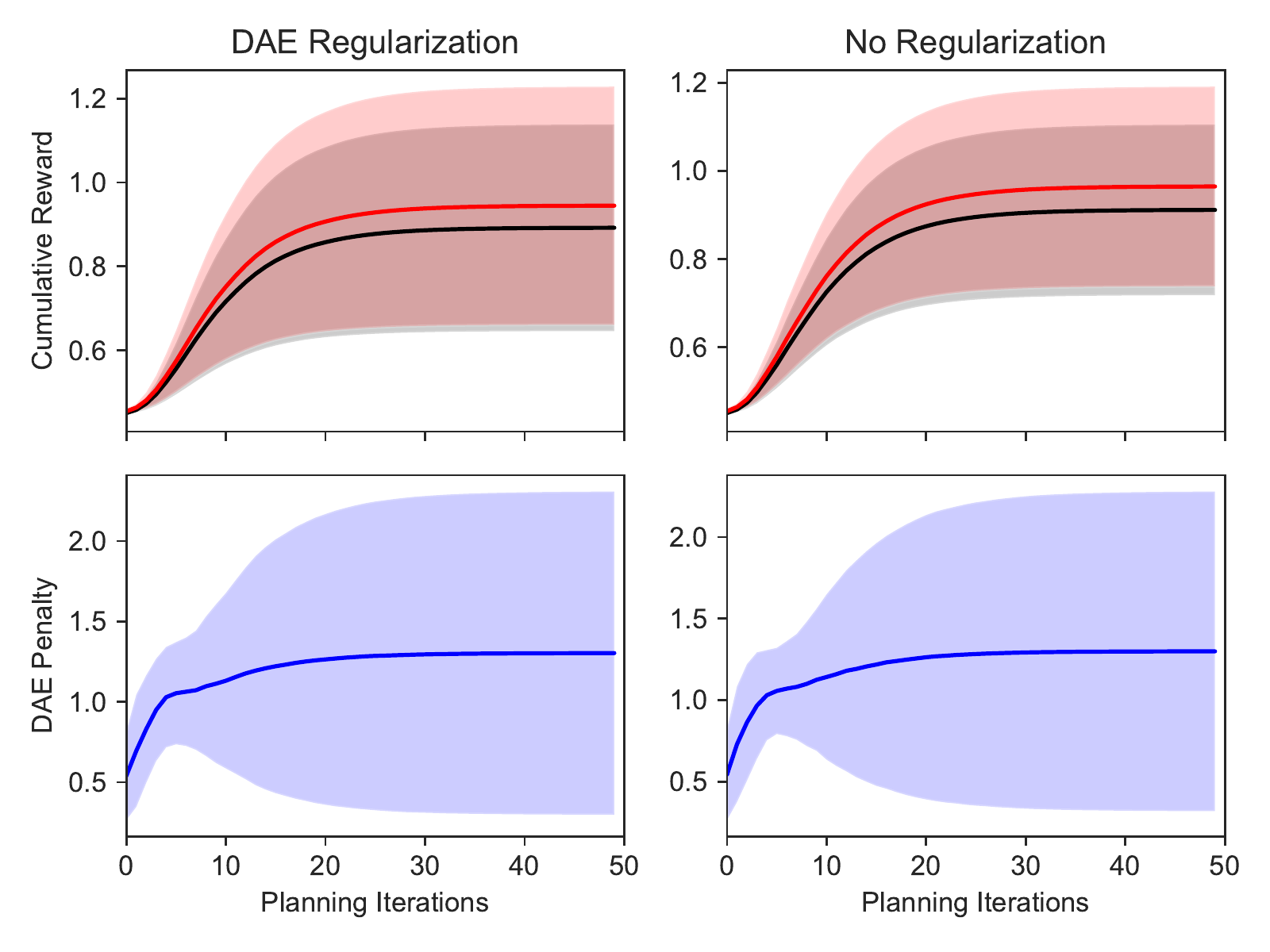}
&
\includegraphics[width=.5\textwidth,trim={3mm 3mm 3mm 3mm},clip]{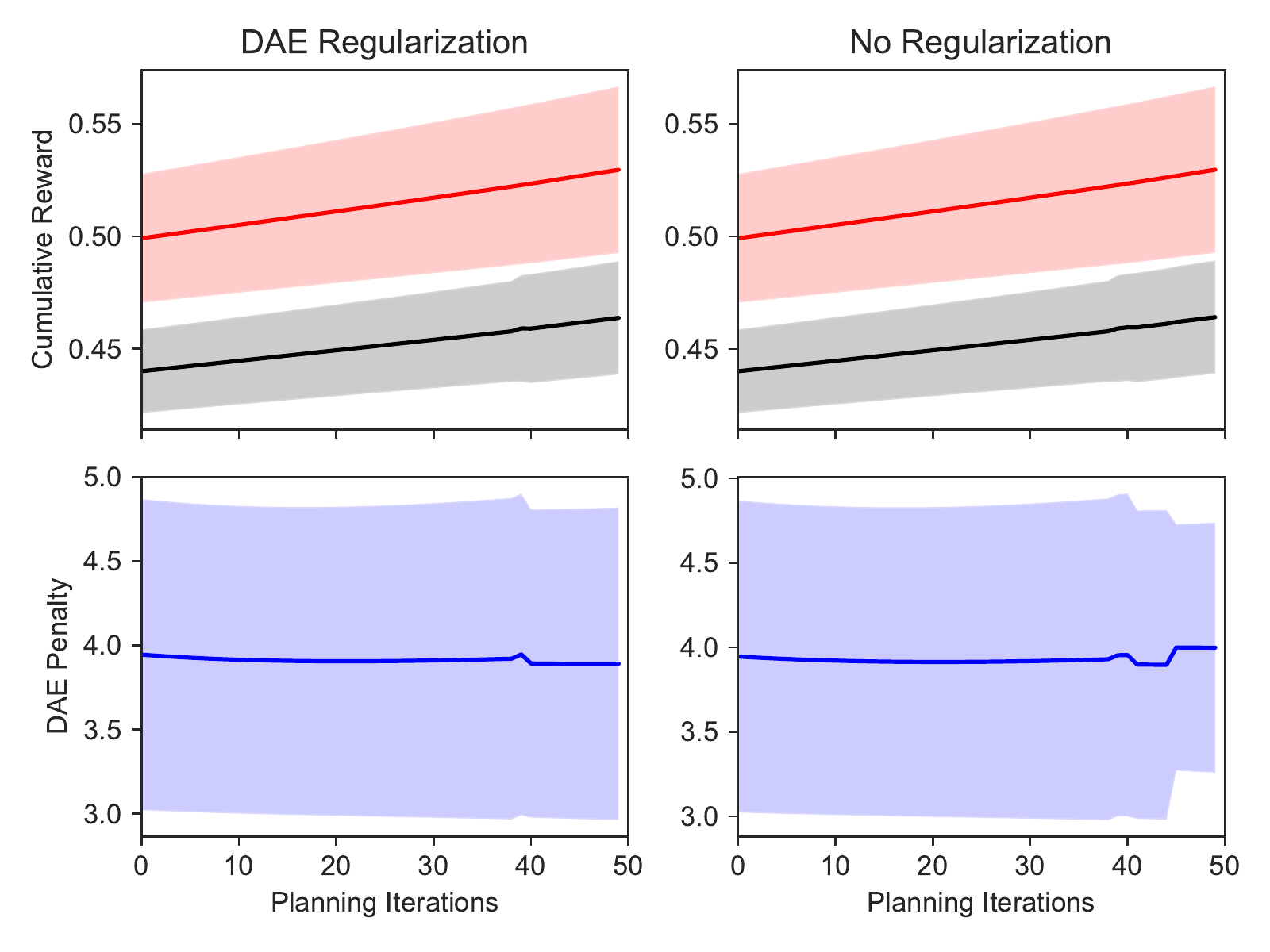}
\\
(a) $t = 0$
&
(b) $t = 0$
\\
\addlinespace[3mm]

\includegraphics[width=.5\textwidth,trim={3mm 3mm 3mm 3mm},clip]{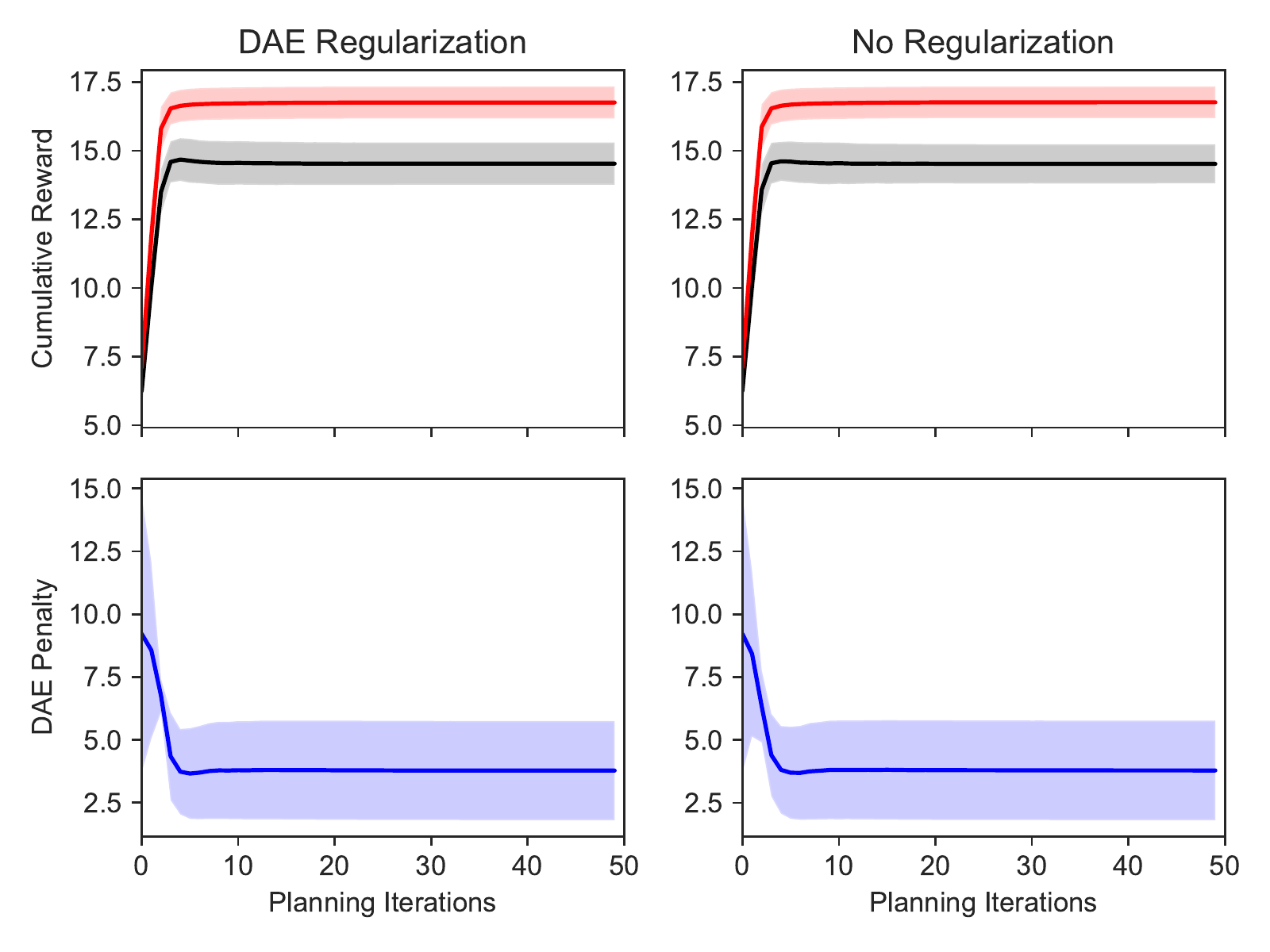}
&
\includegraphics[width=.5\textwidth,trim={3mm 3mm 3mm 3mm},clip]{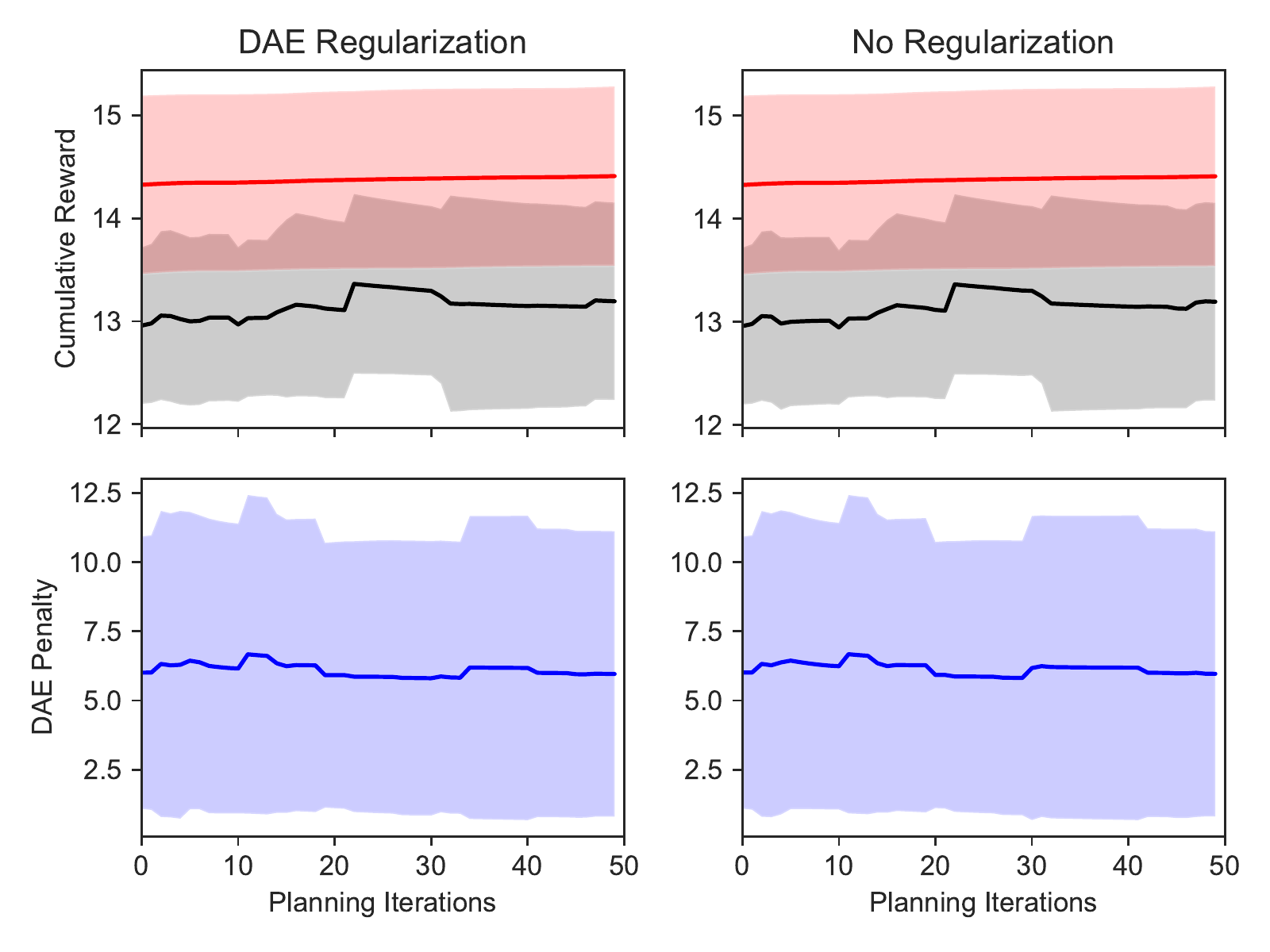}
\\
(c) $t = 10$
&
(d) $t = 10$
\\
\addlinespace[3mm]

\includegraphics[width=.5\textwidth,trim={3mm 3mm 3mm 3mm},clip]{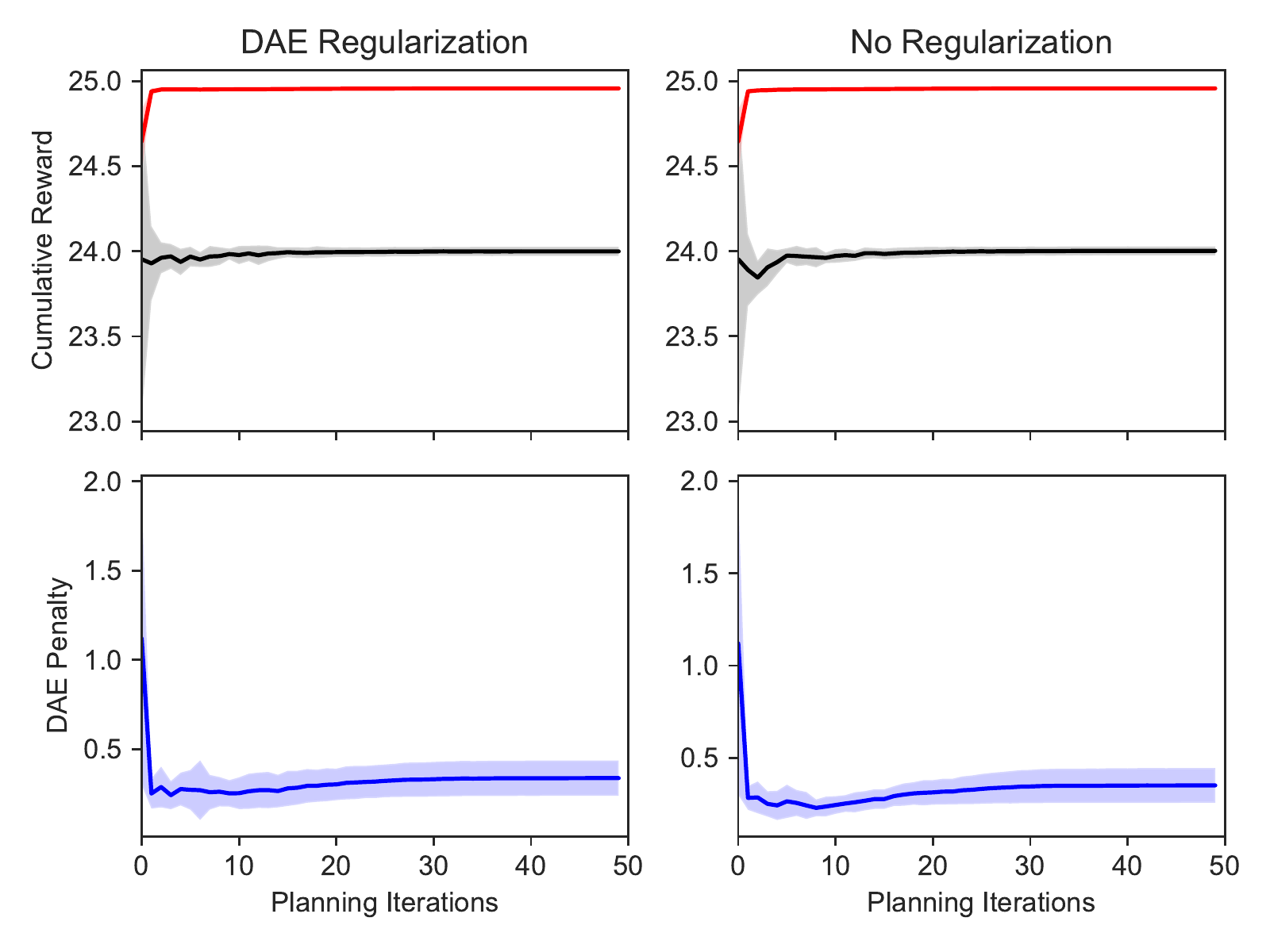}
&
\includegraphics[width=.5\textwidth,trim={3mm 3mm 3mm 3mm},clip]{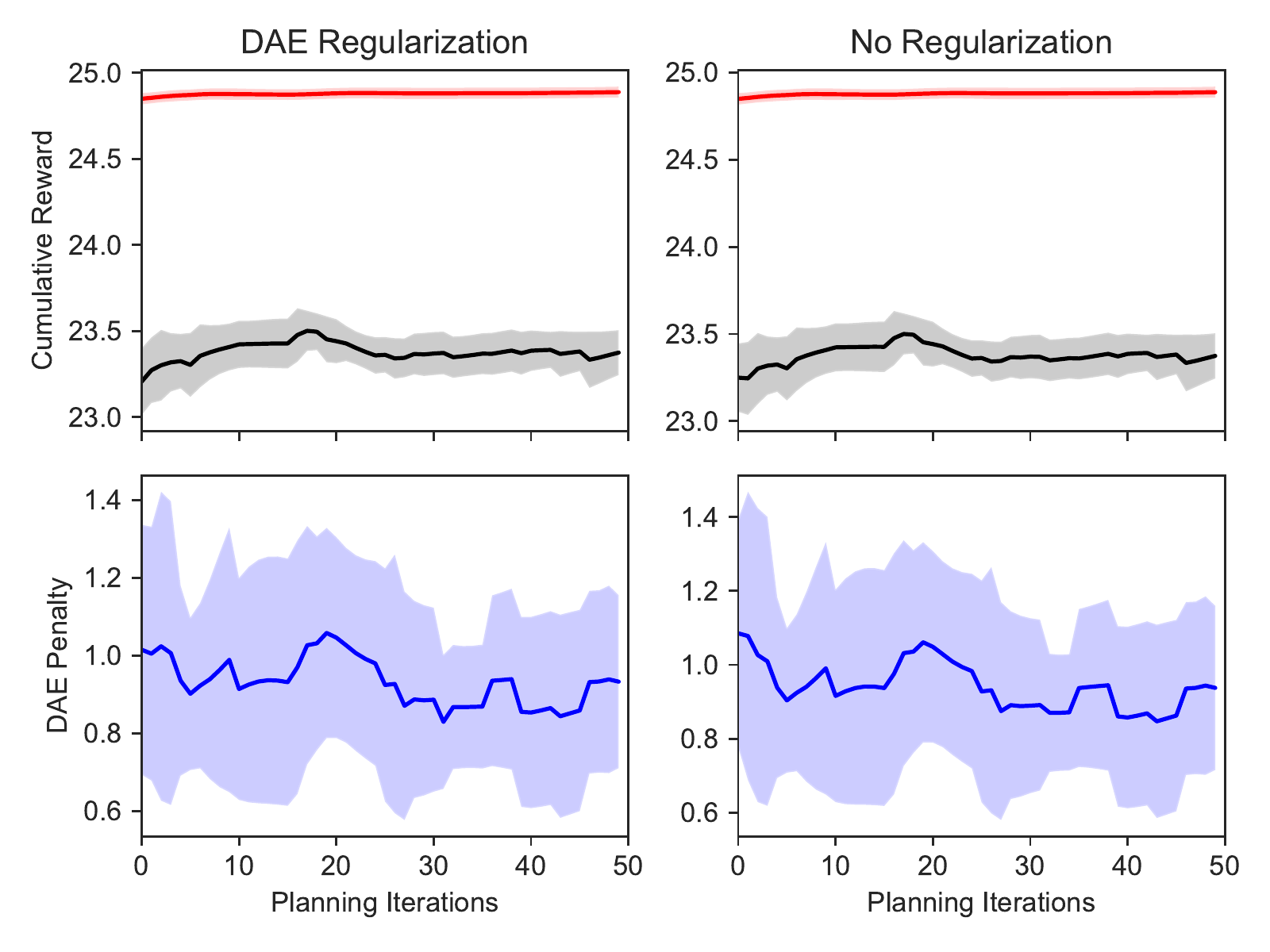}
\\
(e) $t = 50$
&
(f) $t = 50$
\\
\end{tabular}
\end{center}
\caption{Visualization of trajectory optimization at different timesteps $t$ during Episode 5 of end-to-end training in the Cartpole environment. Here, the red line denotes the rewards predicted by the model (imagination) and the black line denotes the true rewards obtained when applying the sequence of optimized actions (reality).}
\label{f:cartpole-traj-opt-figs}
\end{figure*}

\begin{figure*}[tp]
\begin{center}
\begin{tabular}{cc}
Optimizer: CEM & Optimizer: Adam
\\
\addlinespace[3mm]

\includegraphics[width=.5\textwidth,trim={3mm 3mm 3mm 3mm},clip]{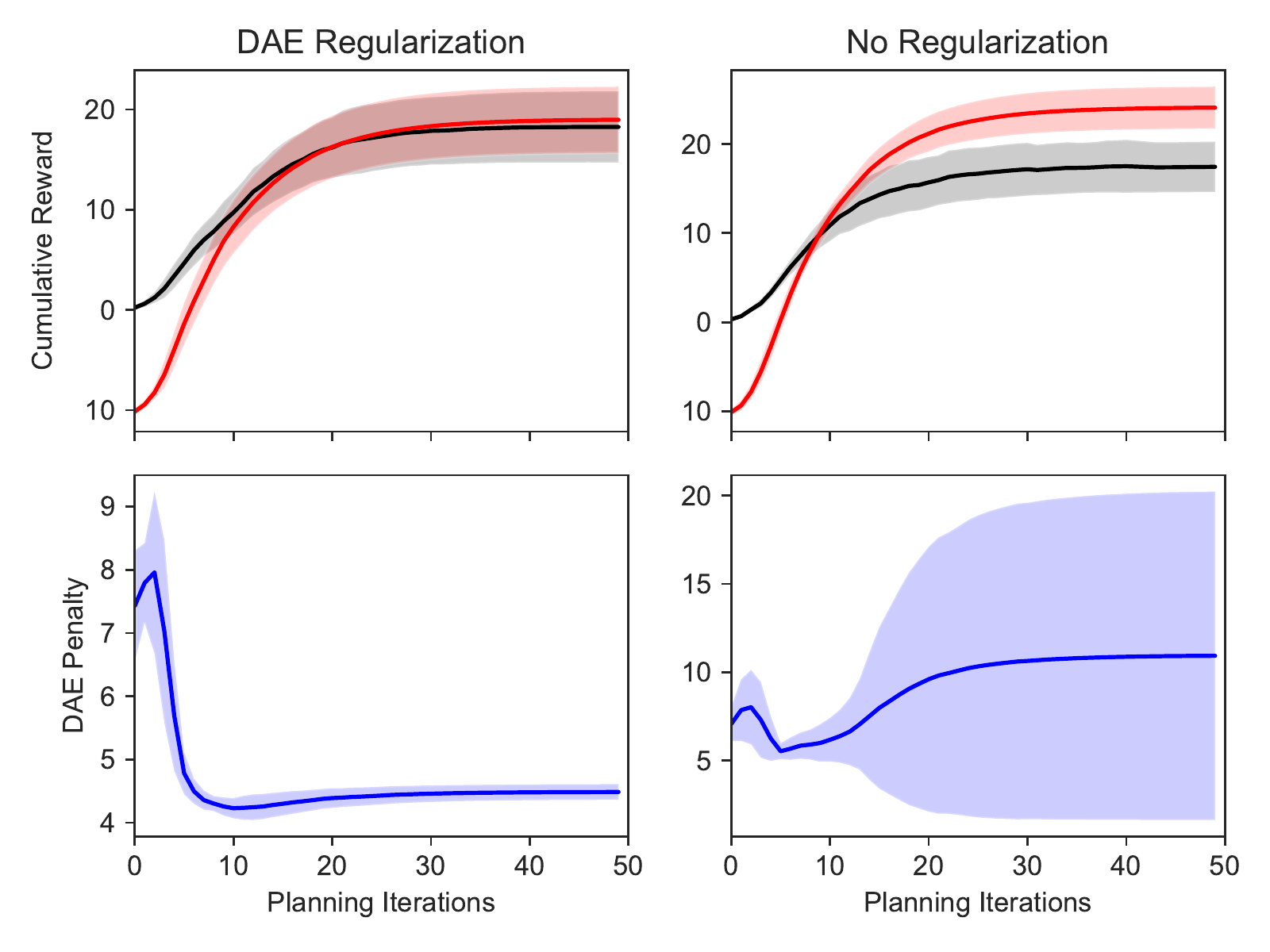}
&
\includegraphics[width=.5\textwidth,trim={3mm 3mm 3mm 3mm},clip]{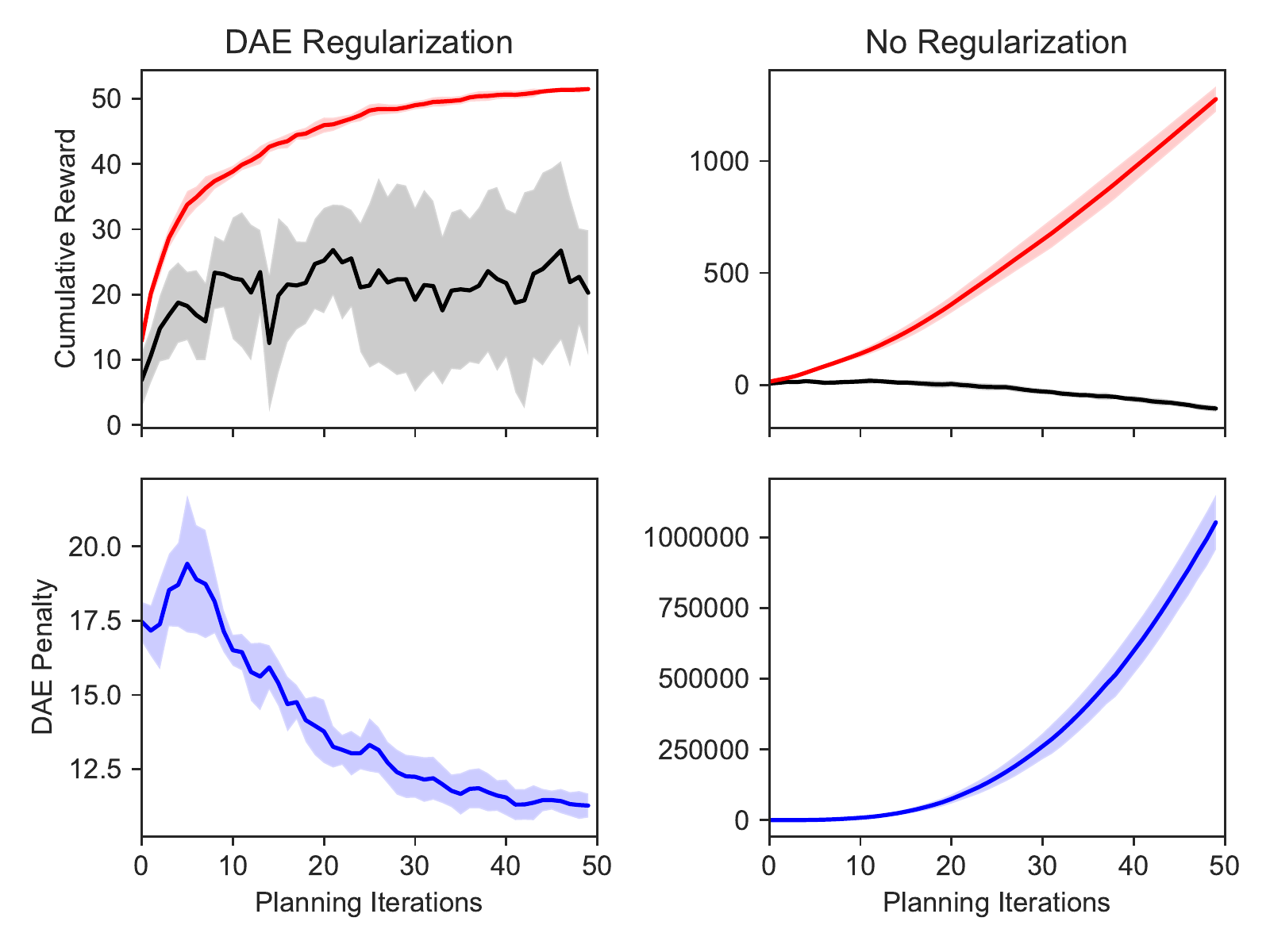}
\\
(a) $t = 0$
&
(b) $t = 0$
\\
\addlinespace[3mm]

\includegraphics[width=.5\textwidth,trim={3mm 3mm 3mm 3mm},clip]{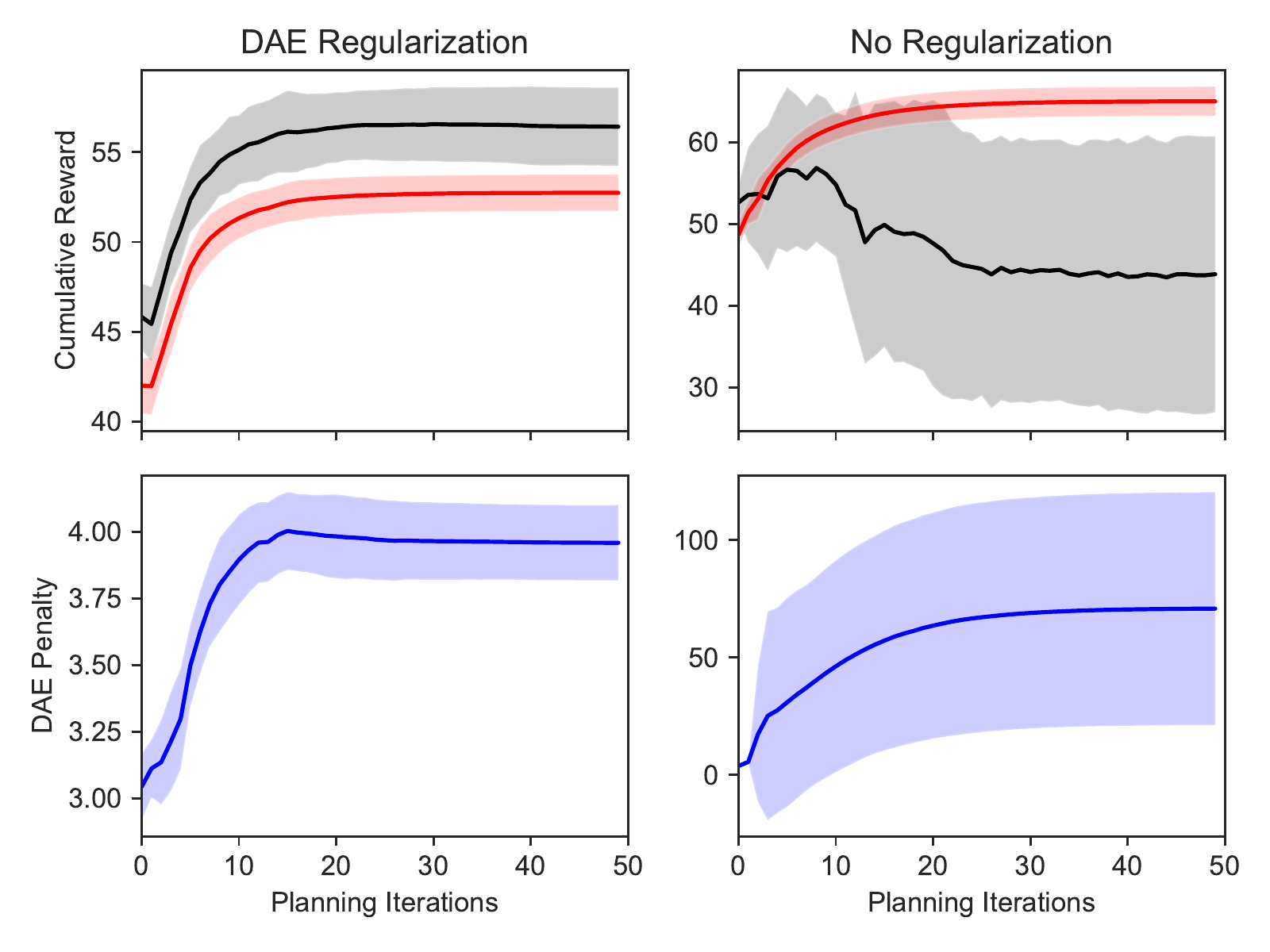}
&
\includegraphics[width=.5\textwidth,trim={3mm 3mm 3mm 3mm},clip]{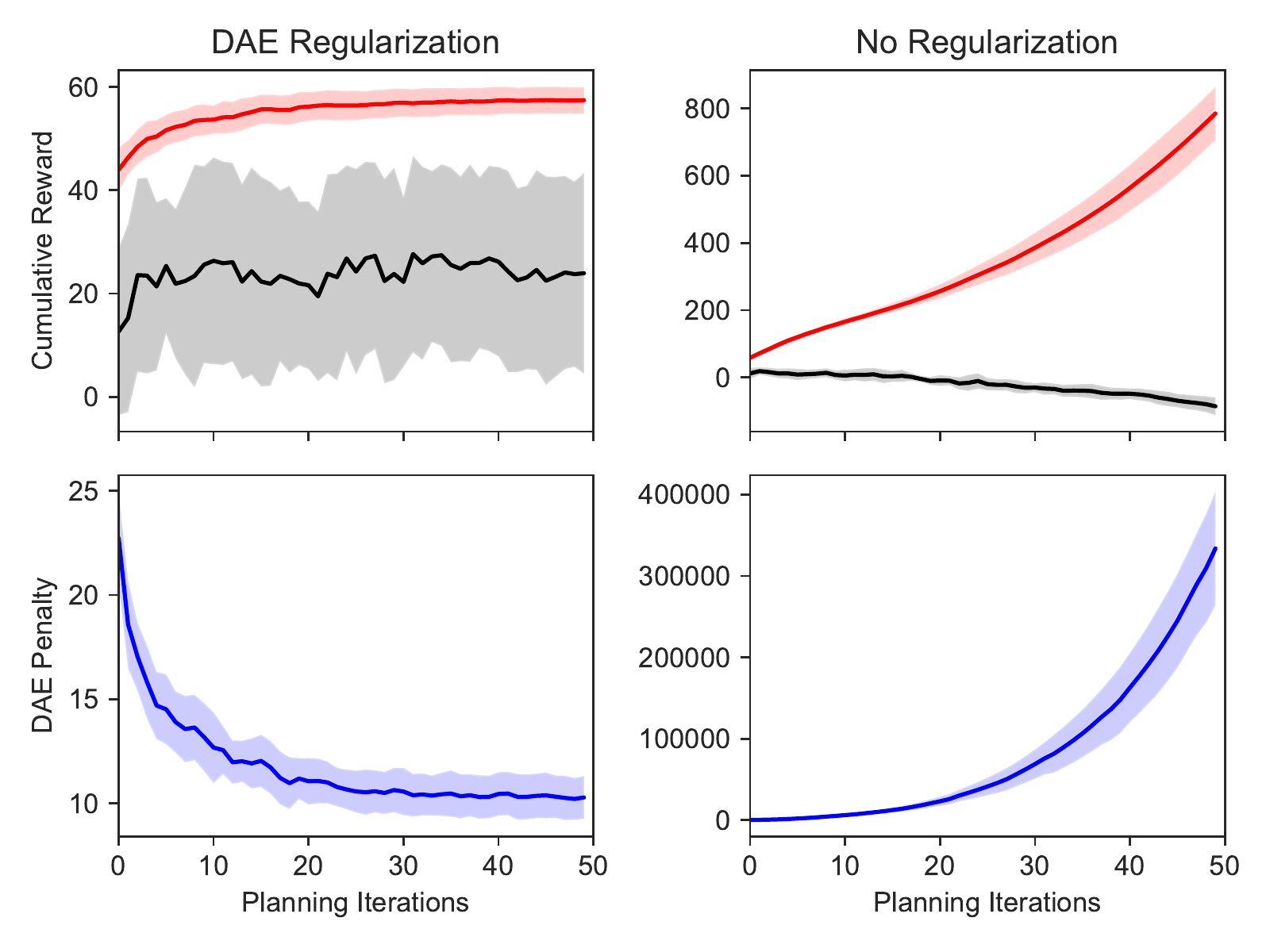}
\\
(c) $t = 10$
&
(d) $t = 10$
\\
\addlinespace[3mm]

\includegraphics[width=.5\textwidth,trim={3mm 3mm 3mm 3mm},clip]{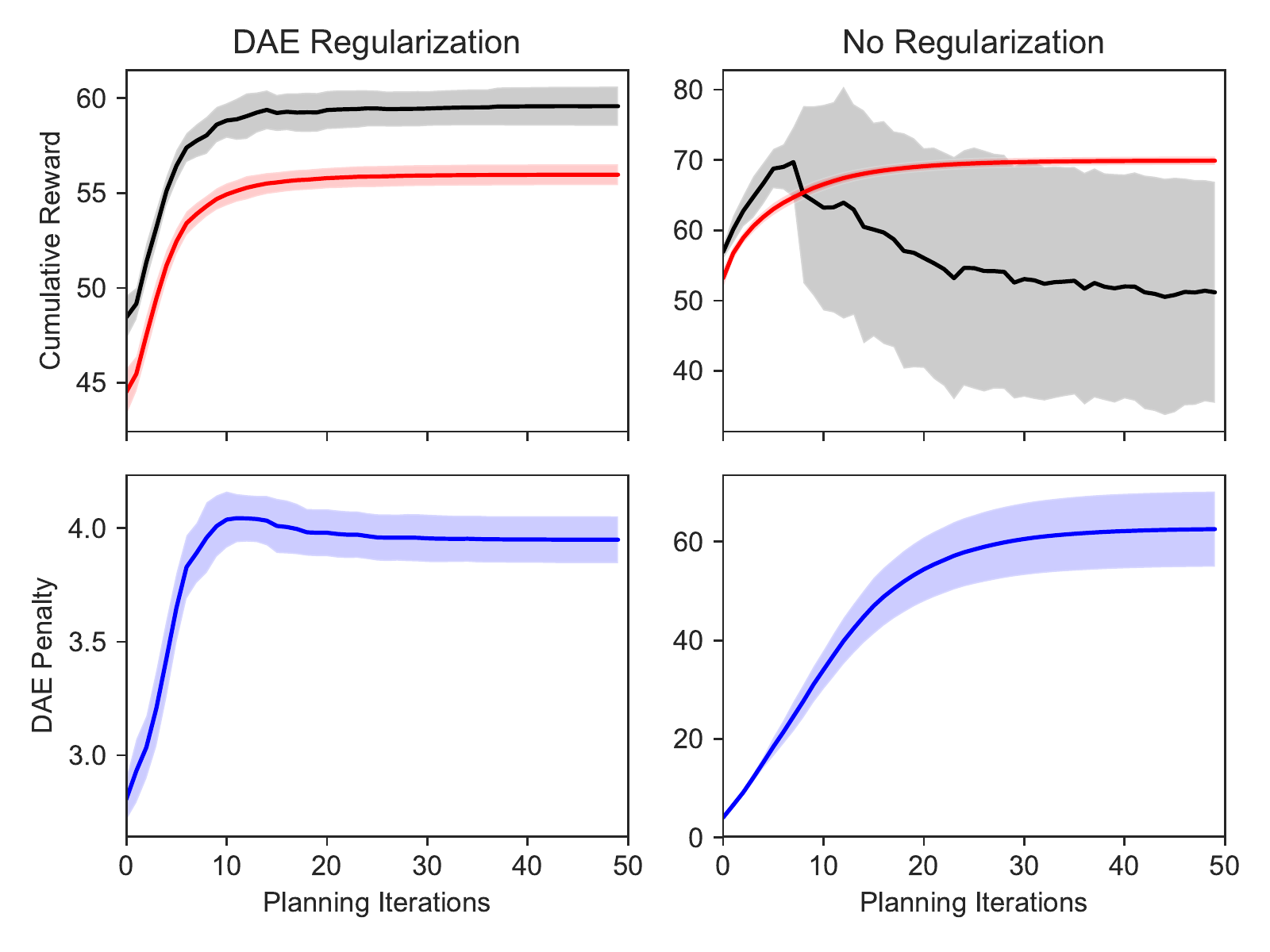}
&
\includegraphics[width=.5\textwidth,trim={3mm 3mm 3mm 3mm},clip]{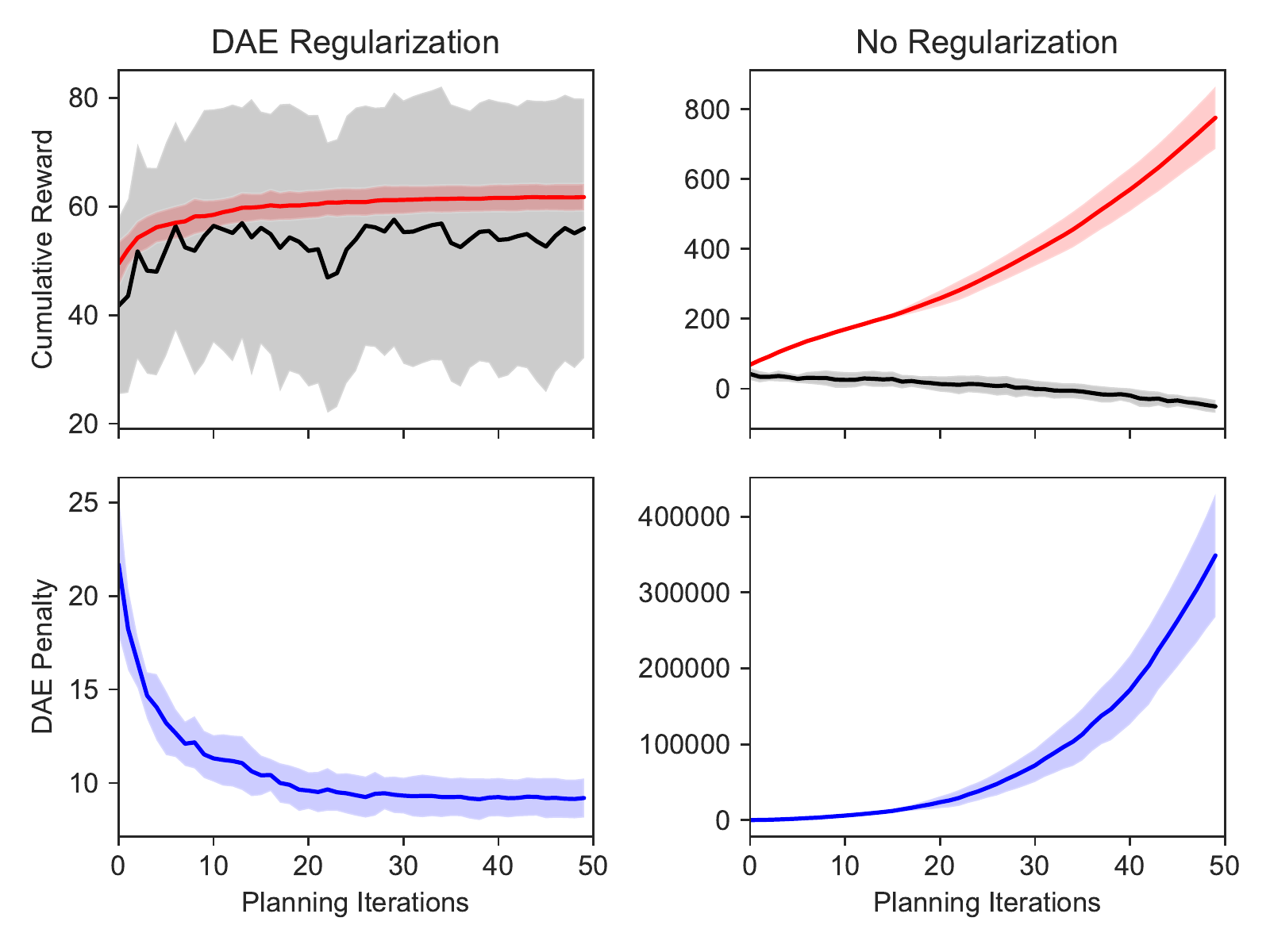}
\\
(e) $t = 50$
&
(f) $t = 50$
\\
\end{tabular}
\end{center}
\caption{Visualization of trajectory optimization at different timesteps $t$ during Episode 5 of end-to-end training in the Half-cheetah environment. Here, the red line denotes the rewards predicted by the model (imagination) and the black line denotes the true rewards obtained when applying the sequence of optimized actions (reality).}
\label{f:hc-traj-opt-figs}
\end{figure*}

In Figures \ref{f:cartpole-traj-opt-figs} and \ref{f:hc-traj-opt-figs}, we visualize trajectory optimization at different timesteps $t$ during Episode 5 of end-to-end experiments in Cartpole and Half-cheetah. It can be observed that the DAE penalty correlates with the inaccuracies of the model and that the DAE regularization is effective in guiding the optimization procedure to remain within the data distribution.

\end{document}